\documentclass{article}

\usepackage{algorithm,algpseudocode,amsmath,amssymb}

 \usepackage[main, final]{neurips_2026}

\usepackage[utf8]{inputenc} 
\usepackage[T1]{fontenc}    
\usepackage{hyperref}       
\usepackage{url}            
\usepackage{booktabs}       
\usepackage{amsfonts}       
\usepackage{nicefrac}       
\usepackage{microtype}      
\usepackage{xcolor}         
\usepackage{amsmath}
\usepackage{amssymb}
\usepackage{enumitem}
\usepackage{multirow} 
\usepackage{booktabs}
\usepackage{graphicx}
\usepackage{amssymb}
\usepackage{booktabs}
\usepackage{multirow}
\usepackage{wrapfig}
\usepackage{subcaption}
\usepackage[table]{xcolor}
\usepackage{pifont}

\usepackage{xcolor}
\usepackage{amssymb}
\newcommand{\ccmark}{%
  \begingroup
  \setlength{\fboxsep}{1pt}%
  {$\checkmark$}%
  \endgroup
}
\usepackage{graphicx}    
\makeatletter
\newcommand\blfootnote[1]{%
  \begingroup
    \renewcommand\thefootnote{}%
    \renewcommand\@makefnmark{}%
    \footnotetext{#1}%
  \endgroup
}
\makeatother
\title{Prism Attention: Proposal-Refined Index Sharing Mechanism for Efficient LLMs Inference}

\author{%
  \normalfont
  Zhenxu Tian\textsuperscript{1,2 $*$}\quad
  Kebin Liu\textsuperscript{2 $*$ $\dagger$}\quad
  Zhengwu Yang\textsuperscript{2}\quad
  Yi Su\textsuperscript{1}\quad
  Qingqing Dang\textsuperscript{2} \\[3pt]
  \normalfont
  Yanlin Sha\textsuperscript{2}\quad
  Kaipeng Deng\textsuperscript{2}\quad
  Dianhai Yu\textsuperscript{2}\quad
  Yanjun Ma\textsuperscript{2 $\ddagger$}\quad
  Juntao Li\textsuperscript{1 $\ddagger$}\quad
  Min Zhang\textsuperscript{1} \\[6pt]
  \normalfont
  \textsuperscript{1}Soochow University \qquad
  \textsuperscript{2}Baidu Inc. \\[3pt]
  \normalfont
  \texttt{zhenxutnlp@outlook.com} \quad \texttt{liukebin@baidu.com}
}

\begin{document}
\maketitle

\blfootnote{$^{*}$Equal contribution.}
\blfootnote{$^{\dagger}$Project lead.\quad $^{\ddagger}$Corresponding author.}
\blfootnote{Work done during an internship at Baidu Inc.}


\begin{abstract}
Long-context decoding in large language models (LLMs) is increasingly bottlenecked by the memory and computation required to attend over an ever-growing Key-Value (KV) cache at each autoregressive step. Although token-level sparse attention reduces this burden by restricting computation to a small subset of relevant past tokens, its practical gains are often offset by repeatedly recomputing layer-wise sparse indices as the context grows. 
A common remedy is to share sparse indices across layers by computing them only at designated anchor layers, but such static reuse ignores the layer-wise evolution of attention patterns, causing the reused indices to progressively deviate from the layer-specific optima and incur substantial attention loss.
By revisiting cross-layer sparsity from the perspective of attention dynamics, we find that \textbf{inter-layer salient token shift is highly localized}: although the exact top-$\kappa$ indices differ across layers, the core high-attention tokens remain stable, with changes concentrated near the expanded boundary of previously selected regions. 
Motivated by this, we propose \textbf{Prism Attention}, a proposal-refined sparse attention framework. It replaces rigid cross-layer index reuse with a coarse-to-fine strategy, preserving layer-wise adaptability while retaining the efficiency benefits of cross-layer sharing.
Extensive experiments show that Prism Attention achieves state-of-the-art performance among mainstream sparse attention methods on both reasoning and long-context benchmarks, while attaining a $3.4\times$ end-to-end decoding speedup over FlashAttention at 128K context length.

\end{abstract}


\begin{wrapfigure}{r}{0.48\textwidth} 
    \vspace{-20pt}
    \centering
    \includegraphics[width=\linewidth]{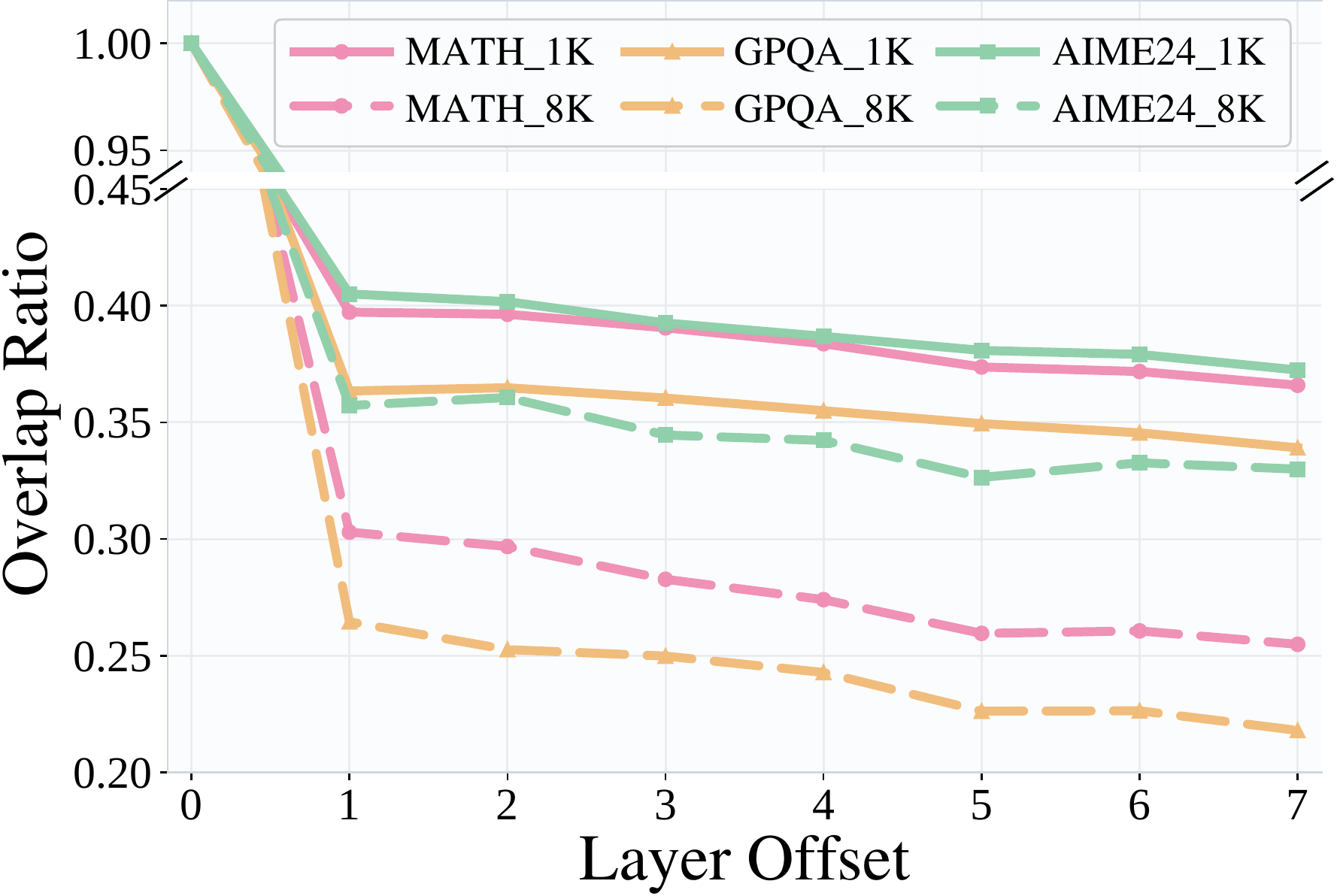}
    \caption{Cross-layer overlap ratio across varying distances of different layer offset.} 
    \label{fig:forward_layer_offsets}
    \vspace{-34pt}
\end{wrapfigure}

\section{Introduction}
\label{sec:intro}
As large language models (LLMs) \cite{claude46,gemini3,guo2025deepseek,openaio1,yang2025qwen3} are pushed toward longer-context reasoning, autoregressive decoding increasingly becomes bottlenecked by attention over the growing Key-Value (KV) cache. 
At each decoding step, the newly generated token must attend over the entire accumulated context, causing the attention cost to scale linearly with sequence length. 
Sparse attention mitigates this by restricting computation to a small subset of relevant past tokens.
Among existing paradigms, token-level \cite{liu2025deepseek,zeng2026glm} sparsity achieves stronger empirical performance than coarse-grained block-wise alternatives \cite{gao2025seerattention,gao2024seerattention,jiang2024minference,lai2025flexprefill,lu2025moba,tang2024quest,tang2026elastic,wang2025proxyattn,xu2025xattention,yuan2025native}. 
However, its practical efficiency is often constrained by sparse index construction: identifying the exact top-$\kappa$ relevant tokens for each query still requires scoring the entire context. As sequence length increases, repeatedly recomputing these layer-wise indices introduces substantial overhead, which can largely offset the theoretical benefits of sparsity.

A natural strategy to circumvent the per-layer selection overhead is cross-layer index sharing, which computes the exact top-$\kappa$ token indices at a few designated anchor layers and reuses them across multiple subsequent layers. While effective in reducing redundant computation, existing approaches \cite{ai2026hylra,bai2026indexcache,deshmukh2025kascade,hao2025omnikv,yang2024tidaldecode,yang2025less} adopt a static sharing paradigm that ignores inter-layer differences by implicitly assuming attention patterns remain invariant with network depth. In reality, attention distributions evolve dynamically across layers. As the distance from the designated anchor layer increases in Figure~\ref{fig:forward_layer_offsets}, the index overlap ratio between the anchor layer and subsequent layers drops sharply and continues to decay. Notably, this degradation accelerates under longer contexts (8K vs. 1K), causing static reuse to repeatedly miss critical tokens and severely degrade long-context performance.

\begin{figure*}
    \centering
    \begin{subfigure}[t]{0.47\textwidth}
        \centering
        \resizebox{\textwidth}{!}{\includegraphics{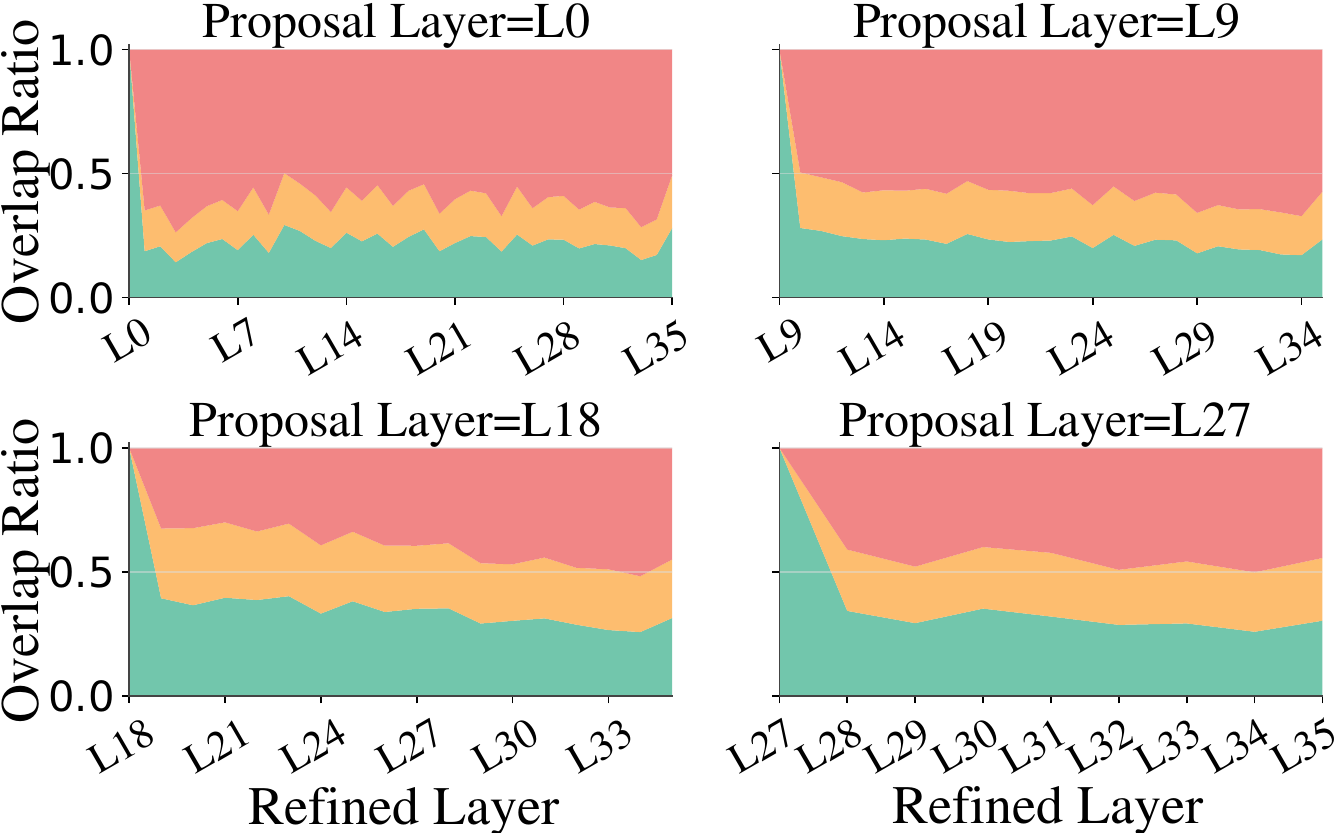}}
         \caption{Layer-wise evolution of refined-layer top-128 indices within expanded proposal regions.}
        \label{fig:Layer_wise_evolution}
    \end{subfigure}
    \hspace{0.5cm}  
    \begin{subfigure}[t]{0.47\textwidth}
        \centering
        \resizebox{\textwidth}{!}{\includegraphics{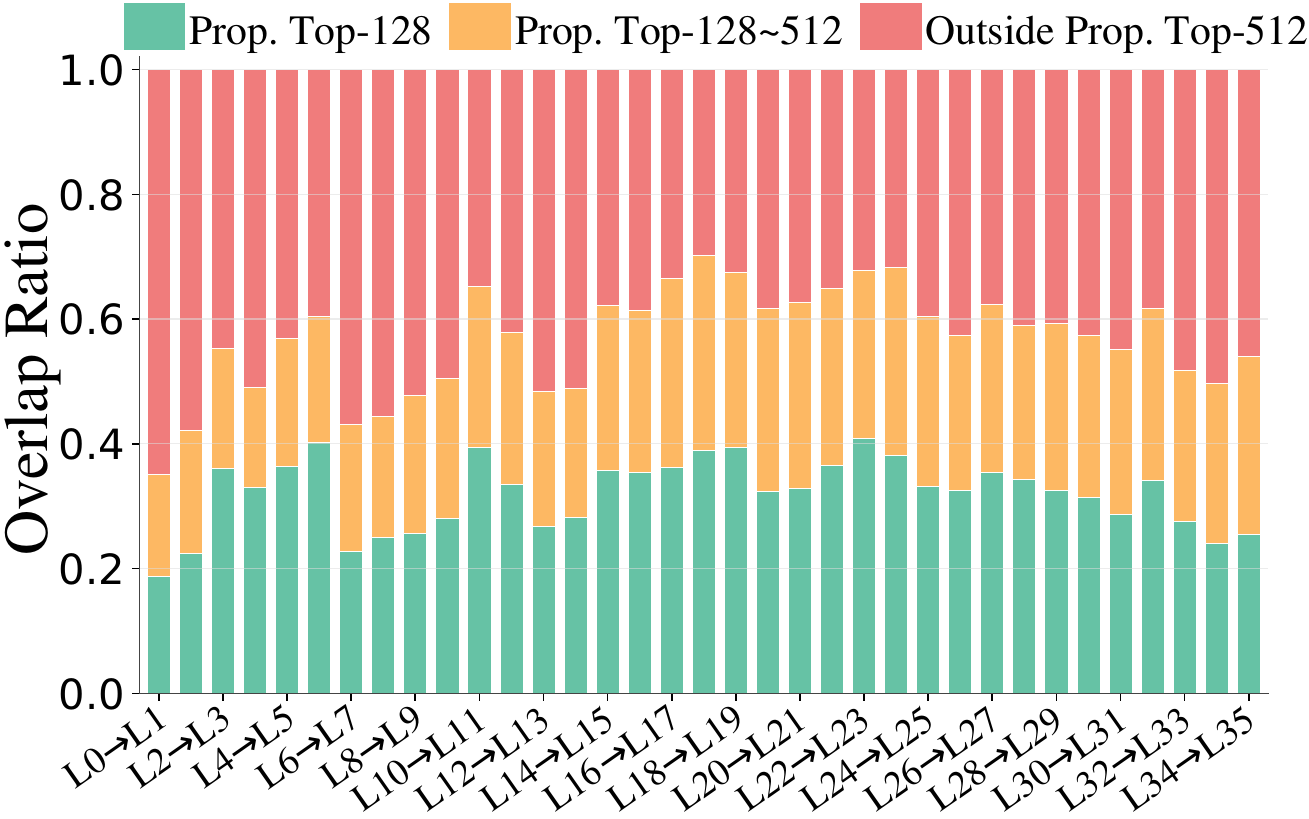}}
        \caption{Adjacent-layer distribution of next-layer top-128 indices within preceding-layer proposal regions.}
        \label{fig:Adjacent_layer_distribution}
    \end{subfigure}
    \caption{\textbf{Localized evolution of sparse indices across layers within proposal regions.}
    For each target (refined) layer, we categorize its own top-128 sparse indices according to their ranks in the corresponding reference (proposal) layer: ``Prop. Top-128'', ``Prop. Top-128$\sim$512'', and ``Outside Prop. Top-512'' denote the ratios of target-layer top-128 indices falling in the reference-layer top-128 set, ranks 128$\sim$511, and outside the top-512 set, respectively. The above two panels share the same legend. (\subref{fig:Layer_wise_evolution}) tracks this distribution from proposal layers $L_0$, $L_9$, $L_{18}$, and $L_{27}$ to their subsequent refined layers, showing that a slightly expanded proposal pool ($\rho=4$, i.e., $4\times$ the base size) consistently captures a significant majority of layer-wise sparse-index variations.
    (\subref{fig:Adjacent_layer_distribution}) measures, for each adjacent pair $L_{\ell}\!\rightarrow\!L_{\ell+1}$, how the top-128 indices of $L_{\ell+1}$ fall into the proposal regions of $L_{\ell}$. This demonstrates that adjacent-layer shifts are heavily concentrated within the expanded proposal window, significantly reducing the risk of missing salient tokens while avoiding the prohibitive cost of full-context search. A more comprehensive analysis is provided in Appendix~\ref{appd_sec:generalization}.}

    \label{fig:localized_evolution}
    \vspace{-20pt}
\end{figure*}

To better understand this limitation, we revisit cross-layer sparsity from the perspective of \emph{attention dynamics}. Our empirical analysis reveals a key property: \textbf{inter-layer salient token shift is highly localized}. As illustrated in Figure~\ref{fig:Layer_wise_evolution} and~\ref{fig:Adjacent_layer_distribution}, although the exact top-$\kappa$ indices differ across layers, these changes are far from arbitrary: the core high-attention tokens remain largely stable, while most changes occur near the boundary of previously selected regions. This indicates that preserving a slightly expanded candidate pool is sufficient to capture layer-wise variations.

Motivated by this observation, we propose \textbf{Prism Attention}, a proposal-refined sparse attention framework for efficient long-context decoding, driven by two parallel algorithmic components: \textbf{Dynamic Index Refinement} and  \textbf{Quantized Proposal Generation}. Instead of rigidly reusing a fixed set of indices, Prism Attention first constructs a relaxed candidate pool at a few designated anchor layers (\emph{Proposal Layer}) and then allows subsequent layers (\emph{Refined Layer}) to perform lightweight token selection within this expanded boundary, thereby adapting to localized inter-layer shifts at low cost. To further reduce the overhead of proposal construction, Prism Attention employs a quantized proposal mechanism that substantially lowers the cost of full-context scoring. Together, these designs enable efficient cross-layer sparsity while preserving attention quality. Extensive experiments show that Prism Attention achieves state-of-the-art performance among mainstream sparse attention methods on both reasoning and long-context benchmarks, while attaining a $3.4\times$ end-to-end decoding speedup over FlashAttention at 128K context length.


Our contributions are as follows:
\begin{itemize}[leftmargin=*, nosep]
\item Revisiting cross-layer sparse attention from the perspective of \emph{attention dynamics} reveals a key property: inter-layer salient token shift is highly localized. This observation reveals the limitation of rigid index sharing and motivates localized cross-layer refinement.
\item We propose \textbf{Prism Attention}, a proposal-refined sparse attention framework for efficient long-context decoding, which replaces static cross-layer index sharing with a coarse-to-fine selection strategy, enabling efficient adaptation to layer-wise attention variation.
\item Two complementary algorithmic components make this design practical: \textbf{Dynamic Index Refinement}, which tracks localized layer-wise token shifts with low overhead, and \textbf{Quantized Proposal Generation}, which reduces the cost of full-context proposal construction via low-bit quantization.

\end{itemize}

\section{Related Work}
\label{related_work}
\subsection{Sparse Attention}
To mitigate the prohibitive quadratic complexity of standard self-attention, particularly during long-context inference, existing sparse mechanisms primarily explore two distinct paradigms: block-level and token-level sparsity. Block-level methods exploit coarse-grained activation patterns and are predominantly tailored to specific inference phases. For instance, works like \cite{fan2026flashprefill,gao2024seerattention,ji2025sale,jiang2024minference,lai2025flexprefill,liu2026rrattention,tang2026elastic,wang2026prism,wang2025proxyattn,xu2025xattention,zhang2025spargeattention} dynamically filter critical blocks to accelerate the computation-heavy prefill phase, whereas approaches such as \cite{gao2025seerattention,sun2025rectified,tang2024quest,yang2025lserve} exclusively optimize the decoding stage by alleviating memory bandwidth bottlenecks associated with KV cache loading. More recently, unified approaches like \cite{hu2025optimizing,lu2025moba,team2025minicpm4,xiao2025optimizing,yuan2025native,zhao2025infllm} have leveraged end-to-end sparse training to accelerate both prefill and decoding phases, while mitigating the accuracy degradation typically associated with block-level sparsity. In contrast, token-level sparsity operates at a strictly finer granularity to preserve critical context that might be discarded by rigid block boundaries. DeepSeek Sparse Attention (DSA) \cite{liu2025deepseek} pioneers the large-scale production deployment of this paradigm, demonstrating its practical efficacy, subsequently validated and adopted by recent models like GLM5 \cite{zeng2026glm}. However, recomputing fine-grained token indices at every layer repeatedly introduces substantial computational overhead, an inefficiency that motivates the exploration of cross-layer redundancy by Prism Attention.

\subsection{Cross-Layer Index Sharing}
Recent studies have revealed significant inter-layer correlation in attention activation patterns, motivating cross-layer index sharing techniques. To exploit this redundancy during sparse decoding, methods such as TidalDecode \cite{yang2024tidaldecode}, LessIsMore \cite{yang2025less}, and OmniKV \cite{hao2025omnikv} directly propagate the top-$\kappa$ indices from designated anchor layers to subsequent layers. To optimize anchor placement, Kascade \cite{deshmukh2025kascade} and HyLRA \cite{ai2026hylra} frame anchor selection as a dynamic programming problem to discover optimal layer configurations. To mitigate the compounding effect of error propagation inherent in this process, IndexCache \cite{bai2026indexcache} employs a greedy search strategy to identify more robust anchor placements. Despite these advances, existing cross-layer index sharing remains constrained by a trade-off between anchor-layer cost and reuse-layer accuracy. Increasing the number of anchor layers provides fresher sparse support but raises index-construction overhead, while sharing indices across more reuse layers reduces overhead but causes reused indices to drift from layer-specific optima.
\section{Observations}
\label{sec:observations}

\begin{figure*}
    \centering
    \begin{subfigure}[t]{0.45\textwidth}
        \centering
        \resizebox{\textwidth}{!}{\includegraphics{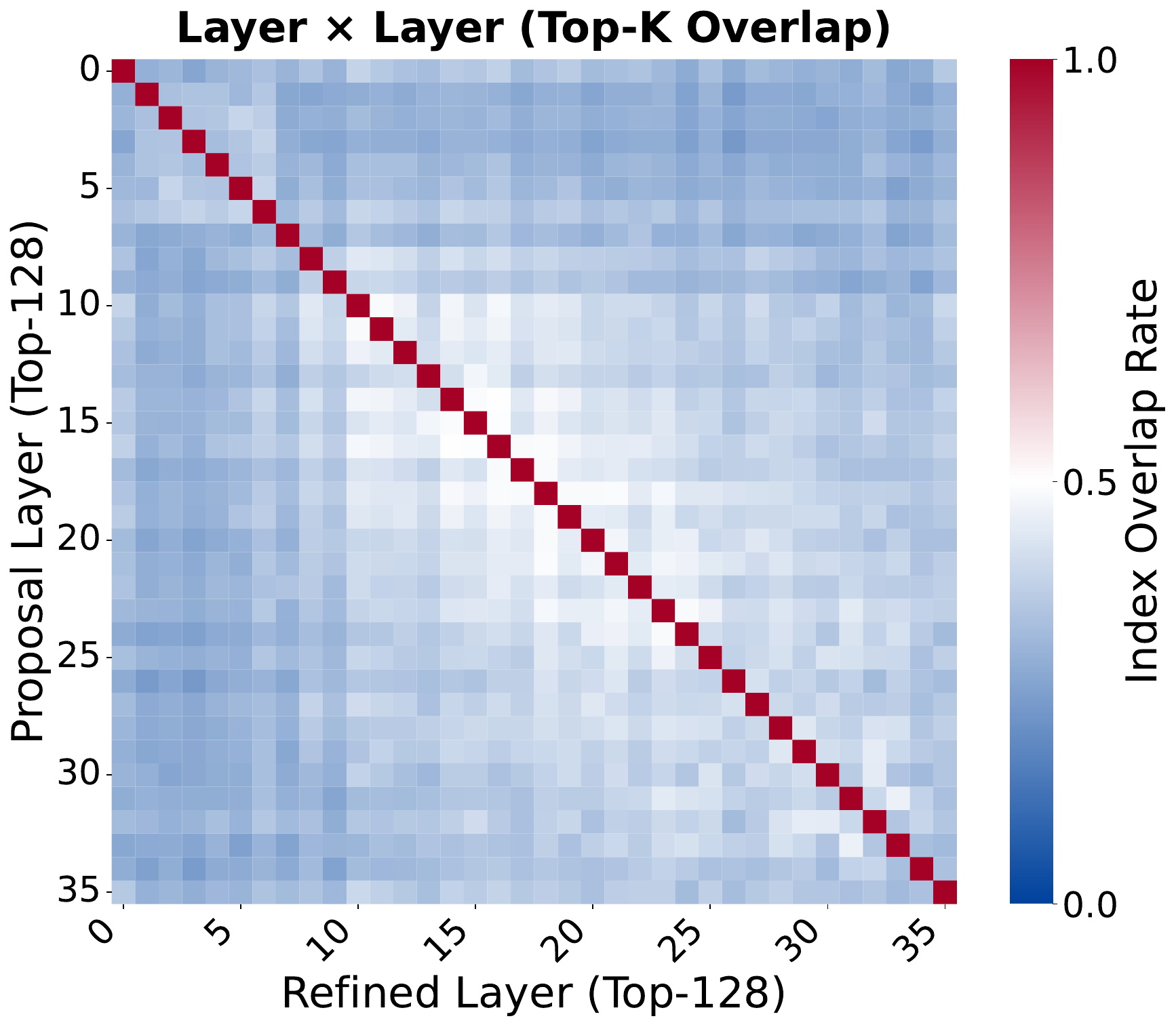}}
    \end{subfigure}
    \hspace{0.5cm}  
    \begin{subfigure}[t]{0.45\textwidth}
        \centering
        \resizebox{\textwidth}{!}{\includegraphics{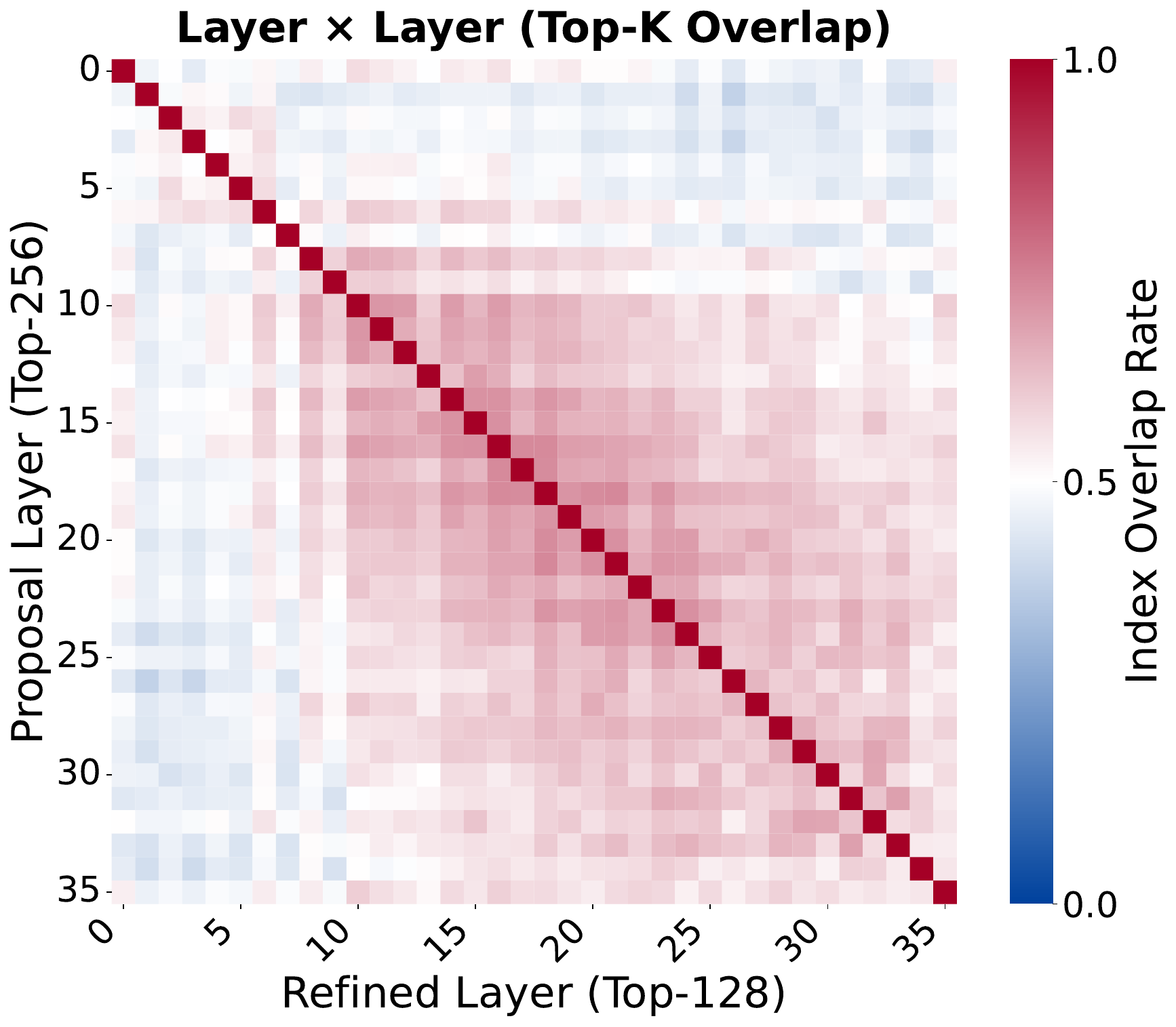}}
    \end{subfigure}
    \caption{
    \textbf{Effect of expanded proposal regions on cross-layer sparse-index coverage.}
    We conduct the analysis on Qwen3-4B using decoding traces collected from AIME24, and report the average cross-layer coverage of refined-layer top-128 sparse indices by the reference-layer proposal set.
    \textbf{Left:} the reference proposal is restricted to its top-128 indices, corresponding to rigid top-$\kappa$ index reuse by the refined layer.
    \textbf{Right:} the reference proposal is expanded from top-128 to top-256 ($\rho=2$), doubling the candidate pool used for refinement while keeping the refined-layer sparse support fixed at top-128.
    Compared with rigid top-128 reuse, the expanded proposal region preserves substantially higher coverage across layers, showing that many refined-layer salient tokens missed by rigid reuse are still captured within a moderately enlarged candidate window.
    }

    \label{fig:index_overlap_heatmap}
    \vspace{-10pt}
\end{figure*}

Motivated by the limitations of static cross-layer sharing discussed in Section \ref{sec:intro}, we conduct an empirical analysis of how sparse attention indices evolve across layers. Figure~\ref{fig:forward_layer_offsets} indicates that, as the distance from the anchor layer increases, the index overlap ratio shifts progressively across depth instead of remaining stationary. As a result, the rigid and non-adaptive nature of static reuse fails to track natural inter-layer drift, causing the reused indices to systematically deviate from the true layer-wise optimum, resulting in severe information loss during autoregressive decoding.

More importantly, we observe that: \textbf{although directly reused indices progressively deviate from the layer-wise optimum, the resulting salient token shift is highly localized rather than arbitrary}. As shown in Figure~\ref{fig:localized_evolution}, high-attention tokens remain stable across adjacent layers, while newly selected tokens beyond the original proposal top-128 are mostly captured by the expanded proposal region between the original and expanded boundaries rather than scattered across the full context. This indicates that most inter-layer variation mainly arises from small local adjustments around previously selected tokens, rather than from a complete reorganization of attention.

This localized structure suggests that the main limitation of static reuse is not the act of reuse itself, but the rigidity of reusing an overly compact top-$\kappa$ set. Since inter-layer variation is concentrated around previously selected tokens, a slightly expanded candidate pool from the anchor layer can already retain high coverage of the salient tokens required by subsequent layers. Figure~\ref{fig:index_overlap_heatmap} further shows that a moderately expanded proposal region can recover much of the lost cross-layer coverage, validating the effectiveness of using an enlarged candidate pool for refinement.




\section{Method}
\label{sec:method}

\begin{figure*}
    \centering
    \includegraphics[width=1\linewidth]{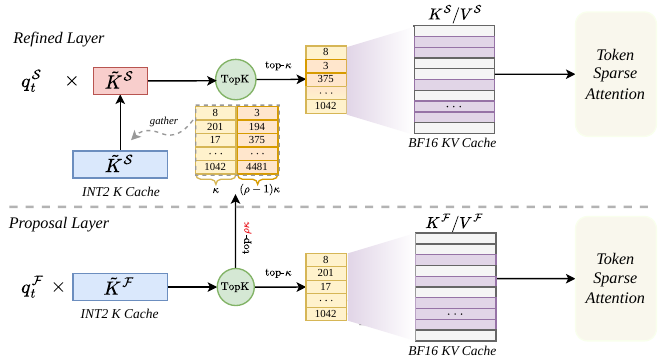}
    \caption{An overview of Prism Attention. Algorithm~\ref{alg:prism} presents the per-step decoding procedure.}
    \label{fig:Prism}
    \vspace{-10pt}
\end{figure*}

\paragraph{Overview.}
Motivated by the localized inter-layer drift observed in Section~\ref{sec:observations},  Prism Attention replaces rigid layer-wise inheritance of a fixed top-$\kappa$ set with a coarse-to-fine cross-layer scheme (Figure~\ref{fig:Prism}): designated anchor layers (\emph{Proposal Layer}) generate an expanded candidate pool, and downstream shared layers (\emph{Refined Layer}) refine their final sparse support within this transmitted proposal region. To make proposal generation itself efficient, Prism further compresses the key cache used at proposal layers through an INT2-based construction scheme. Prism inherits the assignment of Proposal/Refined layers from the original offline IndexCache~\cite{bai2026indexcache} calibration procedure, while redesigning the information transmitted by proposal layers and the way it is consumed by refined layers. Together, these designs make token-level sparse attention adaptive to layer-wise variation while retaining the computational advantages of cross-layer reuse.

\subsection{Proposal-Refined Index Sharing}

Consider autoregressive decoding at step $t$ in transformer layer $\ell$, with query representation $q_t^{\ell}$ and historical key cache $K^{\ell} = \{k_1^{\ell}, k_2^{\ell}, \dots, k_{L_t}^{\ell}\}$, 
where $L_t$ denotes the current context length. Let
\begin{equation}
s_t^{\ell}(i) = \phi\!\left(q_t^{\ell}, k_i^{\ell}\right), \qquad i \in [L_t],
\end{equation}
be the relevance score between the current query and the $i$-th historical key, where $\phi(\cdot,\cdot)$ denotes the sparse scoring function used for token selection, as mentioned in Equation~\ref{eq:qk_max}. The ideal layer-specific sparse support is defined as
\begin{equation}
\mathcal{T}_t^{\ell} = \operatorname{TopK} \big(s_t^{\ell},\kappa\big),
\end{equation}
where $\operatorname{TopK}(\cdot)$ returns the indices of the top-$\kappa$ largest entries. Computing $\mathcal{T}_t^{\ell}$ exactly requires scoring against all $L_t$ cached tokens, incurring $\mathcal{O}(L_t)$ selection cost per layer.

We partition layers into a set of proposal layers and a set of refined layers. For each refined layer $\mathcal{S}$, let $f(\mathcal{S}) \in \mathcal{F}$ denote its associated preceding proposal layer. Conventional static reuse directly copies the selected sparse support from the proposal layer:
\begin{equation}
\mathcal{T}_t^{\mathcal{S}} \leftarrow \mathcal{T}_t^{\mathcal{F}},
\end{equation}
but this strategy assumes that the exact identity of salient tokens remains unchanged across layers. In practice, the layer-wise optimum evolves with depth, and the mismatch between reused indices and true salient tokens accumulates as the distance from the proposal layer increases. 

Prism replaces exact support sharing with \emph{proposal-refined index sharing}. Instead of exporting only a top-$\kappa$ set, the proposal layer $\mathcal{F}$ produces an expanded candidate pool
\begin{equation}
\mathcal{C}_t^{\mathcal{F}} = \operatorname{TopK}\big({s}_t^{\mathcal{F}},n\big), \qquad n = \rho \kappa,\ \rho \ge 1,
\end{equation}
where $\rho$ is a small expansion factor and $n$ is the size of candidate pool. The term ${s}_t^{\mathcal{F}}$ denotes the proposal score used to construct the candidate pool. The proposal layer itself still attends to its own top-$\kappa$ tokens, while caching $\mathcal{C}_t^\mathcal{F}$ for downstream reuse.

Subsequently, for the refined layer $\mathcal{S}$, Prism avoids full-context search and refines the inherited proposal locally. Specifically, it recomputes scores only within $\mathcal{C}_t^\mathcal{F}$:
\begin{equation}
\widehat{s}_t^{\mathcal{S}}(i) = \phi\!\left(q_t^{\mathcal{S}}, k_i^{\mathcal{S}}\right), \qquad i \in \mathcal{C}_t^{\mathcal{F}},
\end{equation}
and then selects its final sparse support as
\begin{equation}
\widehat{\mathcal{T}}_t^{\mathcal{S}} = \operatorname{TopK}\big(\widehat{s}_t^{\mathcal{S}},\kappa\big).
\end{equation}

This design transforms cross-layer reuse from \emph{rigid support sharing} into \emph{proposal-constrained local search}: $\mathcal{C}_t^{\mathcal{F}}$ provides a high-recall candidate region to the layer-wise optimum, and the downstream refinement step absorbs the localized inter-layer deviations.



\subsection{Quantized Proposal Construction}
\label{sec:int2_proposal}

While proposal-refined index sharing restricts full-context scoring to only a few proposal layers, proposal construction remains bandwidth-intensive during long-context decoding, since each proposal layer must repeatedly access a large historical key cache. Prism addresses this bottleneck by compressing the key cache to INT2 for candidate scoring.

Specifically, we quantize historical keys using group-wise asymmetric $b$-bit quantization with $b=2$. To implement this group-wise strategy, we partition the head dimension $d$ of each full-precision key vector $k_i^{{\mathcal{F}}}$ into $M = d/g$ independent sub-vectors of size $g$. The original key vector can thus be denoted as the concatenation of these sub-vectors:
\begin{equation}
k_i^{{\mathcal{F}}} = [k_{i,1}^{{\mathcal{F}}} \parallel k_{i,2}^{{\mathcal{F}}} \parallel \dots \parallel k_{i,M}^{{\mathcal{F}}}]
\end{equation}
where $k_{i,m}^{{\mathcal{F}}} \in \mathbb{R}^g$ represents the $m$-th group. Under the asymmetric quantization scheme, each group $k_{i,m}^{{\mathcal{F}}}$ maintains its own specific scaling factor $\alpha_{i,m}^{{\mathcal{F}}}= (\max(k_{i,m}^{{\mathcal{F}}})-\min(k_{i,m}^{{\mathcal{F}}}))/({2^b-1})$ and zero-point $\beta_{i,m}^{{\mathcal{F}}}=\min(k_{i,m}^{{\mathcal{F}}})$. The quantized 2-bit integer representation $\tilde{k}_{i,m}^{{\mathcal{F}}}$ for the $m$-th group is formally defined as:
\begin{equation}
\tilde{k}_{i,m}^{{\mathcal{F}}} = \text{clip}\left(\left\lfloor \frac{k_{i,m}^{{\mathcal{F}}}-\beta_{i,m}^{{\mathcal{F}}}}{\alpha_{i,m}^{{\mathcal{F}}}} \right \rceil, 0, 2^b - 1 \right)
\end{equation}
where $\lfloor \cdot \rceil$ denotes rounding to the nearest integer. Consequently, the full quantized key vector $\tilde{k}_i^{{\mathcal{F}}}$ is formulated by concatenating these quantized sub-vectors: $\tilde{k}_i^{{\mathcal{F}}} = [\tilde{k}_{i,1}^{{\mathcal{F}}} \parallel \tilde{k}_{i,2}^{{\mathcal{F}}} \parallel \dots \parallel \tilde{k}_{i,M}^{{\mathcal{F}}}]$.

During the autoregressive decoding step $t$, rather than relying on the memory-intensive full-precision cache, we compute approximate proposal scores utilizing the quantized 2-bit keys. Specifically, under the Grouped Query Attention (GQA) architecture, let $q_{t,j}^{{\mathcal{F}}}$ denote the $j$-th query head assigned to a shared KV group $\mathcal{G}$. We define the aggregated proposal score for the $i$-th historical token as:
\begin{equation}
\label{eq:qk_max}
\tilde{s}_t^{{\mathcal{F}}}(i) = \max_{j \in \mathcal{G}} \left\langle q_{t,j}^{{\mathcal{F}}}, \tilde{k}_i^{{\mathcal{F}}} \right\rangle, \qquad i \in [L_t].
\end{equation}
The max-reduction aggregates multiple query heads into one score per KV head, so each KV head is loaded only once per decoding step, avoiding redundant memory accesses across grouped heads. Finally, selecting the top-$\rho\kappa$ scores yields the optimal expanded candidate pool $\mathcal{C}_t^{\mathcal{F}}$.

\section{Experiment}
\label{experiment}

\subsection{Experimental Setup}

\paragraph{Models and Benchmarks.}
Our evaluation is based on Qwen3-4B and Qwen3-8B~\cite{yang2025qwen3}, both of which are specifically trained for reasoning tasks and perform the thinking process by generating extensive tokens. For long-context understanding, we use LongBench~\cite{bai2024longbench} and select the subsets whose average input length exceeds 8192 tokens to focus on genuinely long-input scenarios, capping the maximum generation length at 4096 tokens. For reasoning evaluation, we generate 64, 8, and 16 answers for each problem in the AIME-24/25~\cite{aime}, MATH500~\cite{math500}, GPQA-Diamond~\cite{gpqa} datasets and compute the Pass@1 accuracy, with the maximum generation length set to 32768 tokens.

\paragraph{Baselines.}
We compare Prism Attention against standard full attention and four representative sparse decoding baselines: Quest~\cite{tang2024quest}, LessIsMore (LIM)~\cite{yang2025less}, SeerAttention-R (Seer-R)~\cite{gao2025seerattention}, and IndexCache (IDC)~\cite{bai2026indexcache}. \textbf{Quest} exploits query-aware sparsity by estimating the criticality of each KV cache page from the current query and per-page key extrema, and then attends only to the top-$\kappa$ selected pages. \textbf{LessIsMore} merges per-head importance rankings at a single selection layer into one cross-head token set, which is then unioned with a fixed-ratio recency window and reused by subsequent sparse layers. \textbf{Seer-R} learns block-level attention sparsity through a self-distilled attention gate, enabling query-aware sparse block selection. \textbf{IndexCache} retains indexers only at a few anchor layers whose top-$\kappa$ indices are reused by the consecutive shared layers that follow.

\paragraph{Implementation Details.}
All experiments are conducted on NVIDIA H800 GPUs. For Quest, we use its default block size of 32. For Seer-R, we follow its original configuration, where the block size is set to 64 and all layers perform sparse attention. For IndexCache, we adapt the training-free offline calibration from GLM5~\cite{zeng2026glm} to Qwen3. Specifically, we treat each transformer layer as either an \emph{anchor} layer or a \emph{shared} layer at a 1:3 ratio, and determine the layer assignment by greedy calibration on a held-out subset of OpenR1-MATH-220k~\cite{openr1}. In our implementation, calibration uses 64 samples with a maximum sequence length of 32768 and sets the sparse token budget to $512$. Following the loss-based search strategy of IndexCache, we use token-level cross-entropy as the calibration objective. The resulting partition is fixed for all downstream evaluations. For Prism Attention, we adopt the same offline calibration pipeline as IndexCache to determine the Proposal/Refined partition and set the final sparse budget to $\kappa=2048$, with a proposal expansion factor of $\rho=4$ for long-context tasks ($n=\rho\kappa=8192$) and $\rho=2$ for reasoning tasks ($n=4096$).



\begin{table}[t]
\centering
\small
\caption{Accuracy comparison on reasoning benchmarks using Qwen3-4B and Qwen3-8B.}
\label{tab:reasoning_main}
\begin{tabular}{llccccc}
\toprule
LLMs & Methods & AIME24 & AIME25 & GPQA & MATH500 & Avg. \\
\midrule
\multirow{6}{*}{Qwen3-4B}
& Full Attention & 71.25 & 66.41 & 56.19 & 93.93 & 71.95 \\
& Quest          & 46.67 & 39.79 & 24.74 & 73.40 & 46.15 \\
& Seer-R         & 69.32 & 58.59 & {55.40} & {94.10} & 69.35 \\
& LessIsMore     & 71.48 & 62.87 & \textbf{56.23} & \textbf{94.12} & 71.18 \\
& IndexCache     & 70.99 & 61.30 & 54.29 & 94.05 & 70.16 \\
& Prism          & \textbf{72.60} & \textbf{64.27} & {54.83} & {94.00} & \textbf{71.43} \\
\midrule
\multirow{6}{*}{Qwen3-8B}
& Full Attention & 74.48 & 67.86 & 60.54 & 94.43 & 74.33 \\
& Quest          & 59.79 & 57.71 & 47.22 & 87.80 & 63.13 \\
& Seer-R         & 71.35 & 57.81 & {60.48} & 94.00 & 70.91 \\
& LessIsMore     & 73.00 & 65.24 & \textbf{60.65} & \textbf{94.55} & 73.36 \\
& IndexCache     & 74.84 & 64.47 & 60.39 & {94.40} & 73.53 \\
& Prism          & \textbf{75.05} & \textbf{65.52} & {60.45} & {94.30} & \textbf{73.83} \\
\bottomrule
\end{tabular}

\end{table}

\begin{table*}[t]
\vspace{-8pt}
\caption{Performance comparison of different methods across various LLMs on LongBench.}
\vspace{-7pt}
\centering\small\resizebox{\textwidth}{!}{%
\begin{tabular}{c l cccccccccc}
\toprule
& \textbf{Methods}
& \rotatebox[origin=c]{30}{\textbf{2WikiMQA}}
& \rotatebox[origin=c]{30}{\textbf{TriviaQA}}
& \rotatebox[origin=c]{30}{\textbf{HotpotQA}}
& \rotatebox[origin=c]{30}{\textbf{TREC}}
& \rotatebox[origin=c]{30}{\textbf{Qasper}}
& \rotatebox[origin=c]{30}{\textbf{PCount}}
& \rotatebox[origin=c]{30}{\textbf{MF-en}}
& \rotatebox[origin=c]{30}{\textbf{PRe}}
& \rotatebox[origin=c]{30}{\textbf{GovReport}}
& \textbf{Avg.} \\
\midrule
& Full Attention & 62.68 & 89.69 & 69.79 & 45.00 & 29.35 & 22.26 & 49.16 & 99.06 & 28.66 & 55.07 \\
\multirow{5}{*}[1.1em]{\rotatebox[origin=c]{90}{Qwen3-4B}}
& Seer-R & 59.07 & \textbf{90.42} & 69.59 & 48.00 & 19.19 & \textbf{24.36} & \textbf{50.49} & \textbf{100.00} & 28.19 & 54.37 \\
& LessIsMore & \textbf{62.57} & 82.62 & \textbf{70.73} & 41.00 & 21.43 & 16.67 & 49.83 & 98.00 & 28.42 & 52.36 \\
& IndexCache & 57.73 & 87.28 & 70.64 & 37.00 & \textbf{26.92} & 22.12 & 49.16 & \textbf{100.00} & 28.54 & 53.27 \\
& Prism & 60.36 & 89.76 & 70.16 & \textbf{52.00} & 23.89 & 20.35 & 49.75 & \textbf{100.00} & \textbf{28.57} & \textbf{54.98} \\
\midrule
& Full Attention & 64.77 & 89.02 & 70.45 & 31.33 & 31.52 & 24.95 & 50.54 & 100.00 & 28.66 & 54.58 \\
\multirow{5}{*}[1.1em]{\rotatebox[origin=c]{90}{Qwen3-8B}}
& Seer-R & \textbf{66.17} & 89.59 & 68.58 & 34.00 & 27.53 & 21.67 & 46.51 & \textbf{100.00} & 28.70 & 53.64 \\
& LessIsMore & 58.78 & 89.32 & 68.63 & 31.00 & 24.74 & 21.17 & 37.25 & \textbf{100.00} & 27.93 & 50.98 \\
& IndexCache & 60.50 & \textbf{91.49} & \textbf{71.46} & 27.00 & \textbf{30.58} & 21.19 & 43.45 & 99.12 & \textbf{29.07} & 52.65 \\
& Prism & 65.19 & 91.19 & 70.61 & \textbf{34.03} & 25.67 & \textbf{27.11} & \textbf{52.12} & \textbf{100.00} & 28.81 & \textbf{54.97} \\
\bottomrule
\end{tabular}
}
\label{table:longbench_prism}
\vspace{-17pt}
\end{table*}

\subsection{Accuracy Results} 

\begin{wrapfigure}{r}{0.505\textwidth}
    \centering
    \vspace{-20pt}
    \resizebox{\linewidth}{!}{\includegraphics{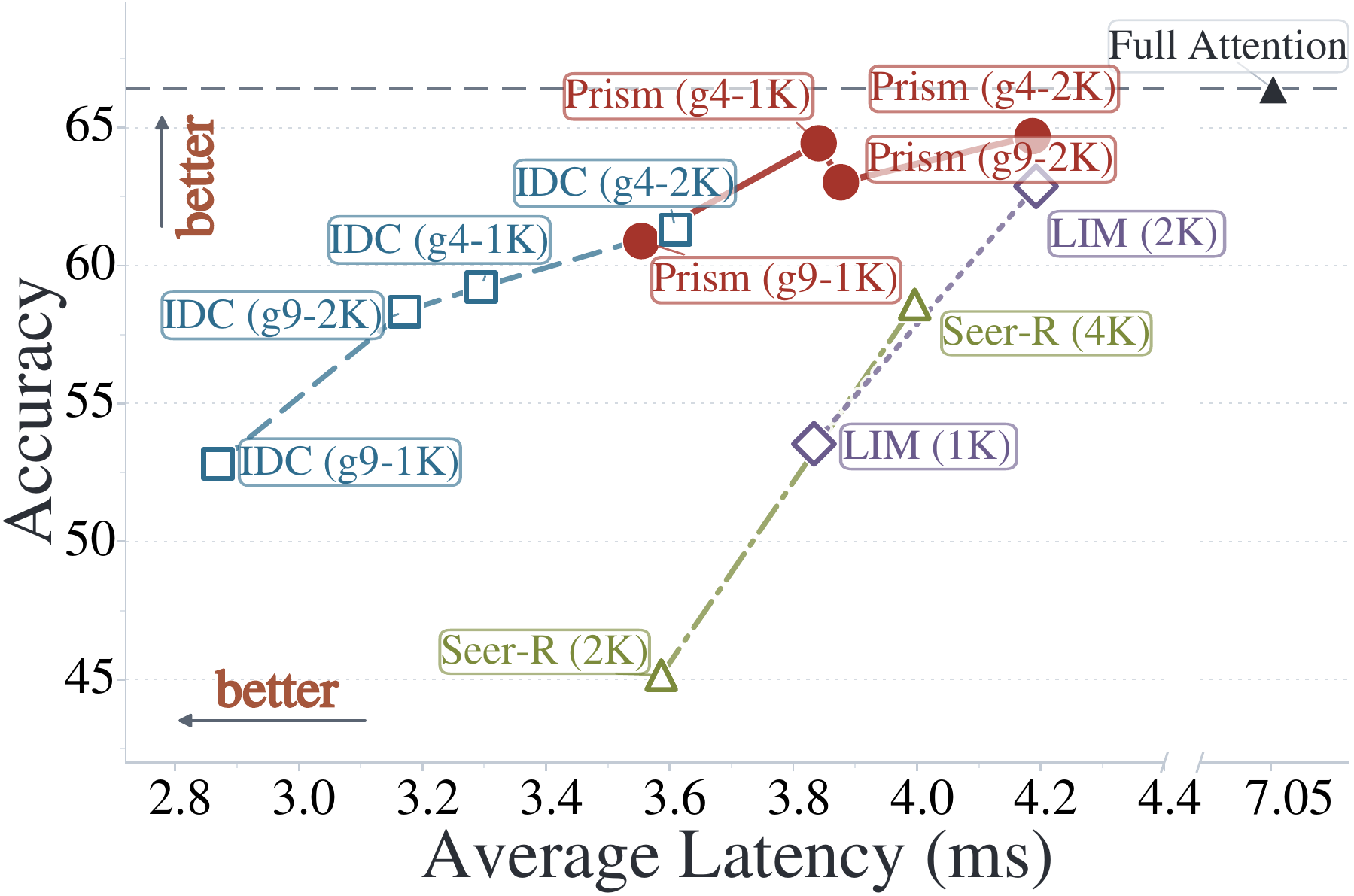}}
    \vspace{-15pt}
    \caption{Accuracy-Efficiency Pareto Comparison.}
    \label{fig:pareto}
    \vspace{-15pt}
\end{wrapfigure}

\paragraph{Performance on Reasoning Benchmarks.}
As shown in Table~\ref{tab:reasoning_main}, Prism achieves the best average accuracy among all evaluated sparse baselines on both Qwen3 models. On Qwen3-4B, Prism reaches an average score of 71.43, improving over Quest, Seer-R, and IndexCache by 25.28, 2.08, and 1.27 points, respectively. This advantage transfers to Qwen3-8B, where Prism narrows the gap with Full Attention to 0.50 points (73.83 vs. 74.33). Notably, the improvements are most pronounced on the two most challenging subsets: AIME24 and AIME25, where the average gain reaches 1.46 points over IndexCache and 1.21 points over LessIsMore. These show that proposal-refined index sharing improves performance on difficult long-reasoning tasks, without degrading general reasoning accuracy.


\paragraph{Accuracy on Long-Context Understanding.}
 As reported in Table~\ref{table:longbench_prism}, Prism achieves the strongest average performance among all sparse attention methods. On Qwen3-4B, Prism obtains an average score of 54.98, nearly matching Full Attention and improving over IndexCache by 1.71 points. On Qwen3-8B, Prism reaches 54.97, slightly exceeding Full Attention and improving over Seer-R and IndexCache by 1.33 and 2.32 points, respectively. These results support our hypothesis that rigid cross-layer index reuse accumulates attention loss as reused indices drift from layer-wise optima, whereas Prism allows subsequent refined layers to recover their own salient tokens within the proposal region, retaining task-relevant context that a compact top-$\kappa$ set would miss.

\begin{figure*}
    \centering
    \begin{subfigure}[t]{0.48\textwidth}
        \centering
        \resizebox{\textwidth}{!}{\includegraphics{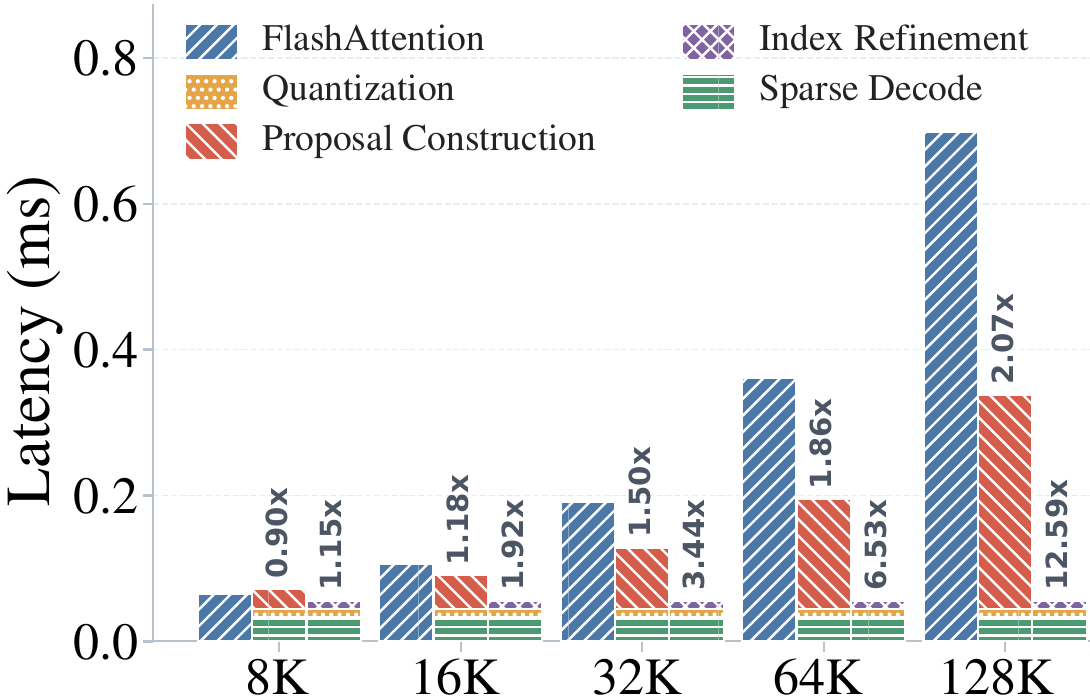}}
        \caption{}
        \label{fig:efficency_latency}
    \end{subfigure}
    \hfill
    \begin{subfigure}[t]{0.48\textwidth}
        \centering
        \resizebox{\textwidth}{!}{\includegraphics{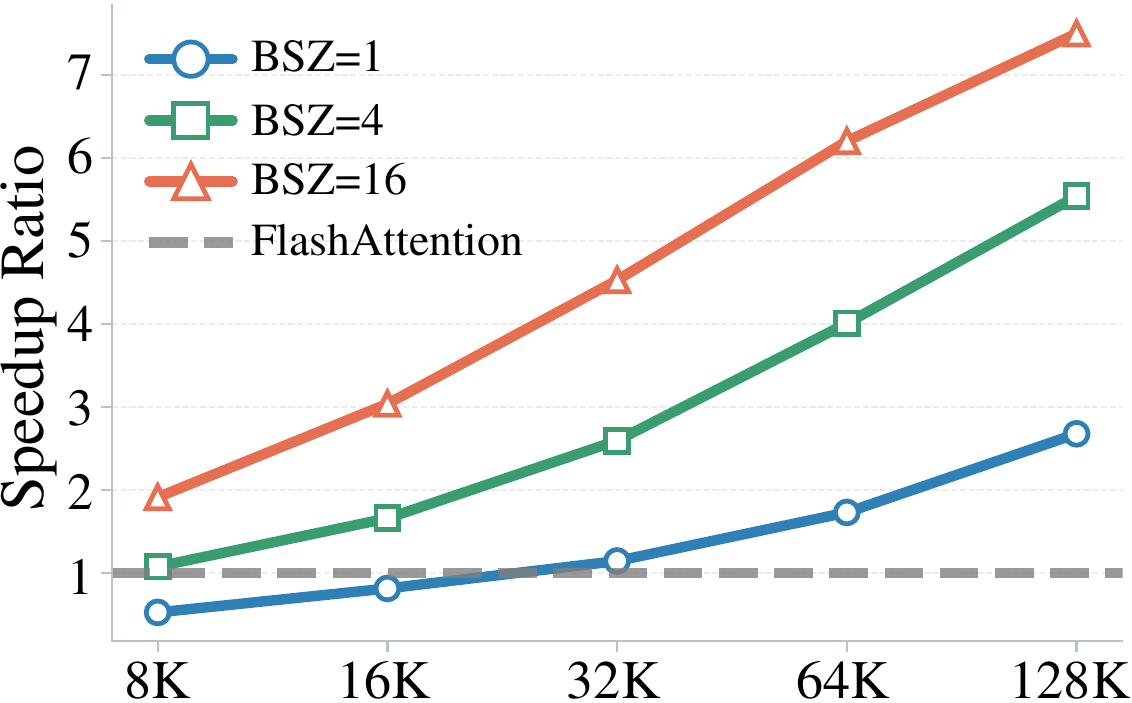}}
        \caption{}
        \label{fig:efficency_speedup}
    \end{subfigure}
    \vspace{-7pt}
    \caption{(\subref{fig:efficency_latency}) Kernel latency breakdown at BSZ=4 across different context lengths. (\subref{fig:efficency_speedup}) The speedup of the average per-layer kernel latency over FlashAttention across context lengths and batch sizes.}
    \vspace{-8pt}
\end{figure*}

\subsection{Efficiency Analysis}

We analyze efficiency at three levels: end-to-end serving latency, attention
kernel latency, and the resulting accuracy-efficiency trade-off.

\begin{wrapfigure}{r}{0.50\textwidth}
    \centering
    \resizebox{\linewidth}{!}{\includegraphics{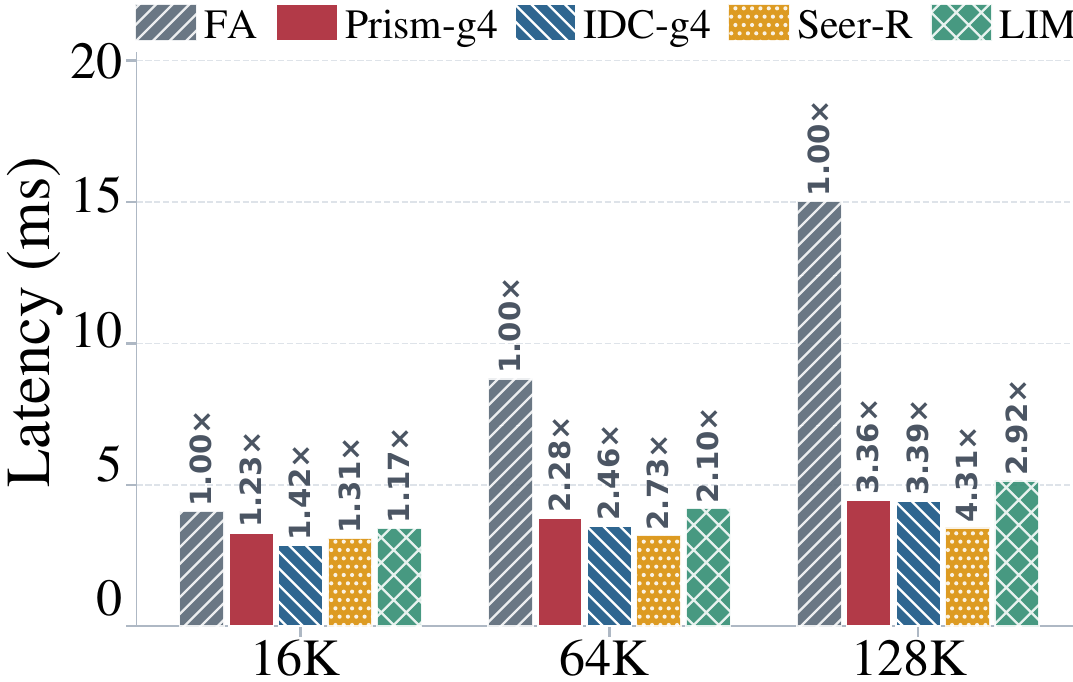}}
    \vspace{-10pt}
    \caption{End-to-end latency comparison across different context lengths on Qwen3-4B at BSZ=16.}
    \label{fig:end_to_end_latency_advanced_bsz16_new}
    \vspace{-13pt}
\end{wrapfigure}

\paragraph{End-to-end latency.} We measure end-to-end decoding latency on Qwen3-4B with SGLang under a sparse budget of 2048. As shown in Figure~\ref{fig:end_to_end_latency_advanced_bsz16_new}, Prism reduces the per-step latency of FlashAttention from 15.03\,ms to 4.47\,ms at 128K (BSZ=16), a $3.36\times$ speedup that increases further with context length. More importantly, the latency overhead of Prism-g4 over IndexCache-g4---incurred by the additional refinement at every refined layer---decreases monotonically as the context grows: Prism is 15.7\% slower at 16K, 8.0\% at 64K, and only 0.7\% at 128K (4.467\,ms vs.\ 4.434\,ms). Refinement scans only the inherited candidate pool, so its cost is independent of context length. These results demonstrate that the lightweight refinement overhead is increasingly amortized in long-context serving, enabling substantially stronger layer-wise adaptability with only marginal additional latency.

\paragraph{Kernel-level analysis.} We evaluate the efficiency of Prism using our customized Triton~\cite{tillet2019triton} decoding kernels by CUDA graph execution time ($\kappa=1024$, $\rho=4$). For proposal layers, the latency consists of key-cache 2bit quantization, proposal construction, and sparse decoding, i.e.,
$$
T_{\text{proposal}}=T_{\text{INT2Decode}}+T_{\text{ProposalConstruction}}+T_{\text{SparseDecode}},
$$
while refined layers replace full proposal construction with lightweight local
index refinement:
$$
T_{\text{refine}}=T_{\text{INT2Decode}}+T_{\text{IndexRefinement}}+T_{\text{SparseDecode}},
$$
giving an effective per-layer latency of
$T=(T_{\text{proposal}}+3T_{\text{refine}})/4$ under the calibrated 1:3 ratio.
Figure~\ref{fig:efficency_latency} shows that the two layer types differ almost only in the index stage: at BSZ=4, proposal construction grows linearly with context, while index refinement remains constant at about 0.011 ms, as do INT2 Quantization and sparse decoding. A proposal layer is $2.07\times$ faster than FlashAttention at 128K whereas a refined layer reaches $12.59\times$, with a latency that is independent of context length. Since three out of every four layers are refined, the costly proposal construction is amortized in the per-layer latency $T$: Figure~\ref{fig:efficency_speedup} shows that its speedup over FlashAttention
grows with both context length and batch size, reaching $7.50\times$ at 128K
with BSZ=16. 

\paragraph{Accuracy-efficiency trade-off.} We provide an end-to-end accuracy-efficiency Pareto comparison using SGLang, where each point represents the average per-token decoding latency over the corresponding average generation length. We evaluate Prism against representative sparse baselines on AIME 25 with Qwen3-4B under different sparse budgets. The notation g$x$ denotes that one in every $x$ layers acts as a proposal layer, while 1K/2K denote the final sparse budget $\kappa$. As shown in Figure~\ref{fig:pareto}, Prism occupies the high-accuracy end of the Pareto frontier: Prism-g4-1K reaches 64.43 accuracy at 3.84\,ms, $1.84\times$ faster than Full Attention while retaining 97.0\% of its accuracy, and strictly dominates both LIM-2K (62.87 at 4.19\,ms) and Seer-R-4K (58.59 at 4.00\,ms); IDC-g4-1K is 0.55\,ms faster but 5.21 points less accurate. The results show that Prism achieves a favorable accuracy-efficiency trade-off across different sparse budgets and Proposal/Refined configurations.

\subsection{Ablation}
\label{subsec:ablation}

\begin{table*}[t]
\caption{Ablation study of index construction and proposal-refined sharing on Qwen3-4B.  Group $x$ denotes that one in every $x$ layers acts as a proposal layer (i.e., a proposal-to-refined ratio of $1{:}(x-1)$). All configurations use a sparse budget of $\kappa=1024$; in proposal-refined configurations, each refined layer selects its final $\kappa$ tokens from a candidate pool of $n=\rho\kappa$ with $\rho=4$.}
\centering
\scriptsize
\setlength{\tabcolsep}{3pt}
\renewcommand{\arraystretch}{1.05}
\resizebox{0.95\textwidth}{!}{%
\begin{tabular}{c c c c c c cc}
\toprule
\multirow{2}{*}{\textbf{Per-Layer Indices}} &
\multirow{2}{*}{\textbf{Reuse Indices}} &
\multirow{2}{*}{\textbf{Quantization}} &
\multirow{2}{*}{\textbf{Group}} &
\multirow{2}{*}{\textbf{Calibration}} &
\multirow{2}{*}{\textbf{Proposal-Refined}} &
\multicolumn{2}{c}{\textbf{Qwen3-4B}} \\
\cmidrule(lr){7-8}
& & & & & & \textbf{AIME24} & \textbf{AIME25} \\
\midrule
\ccmark &  &  &  &  &  & 71.35 & 63.87  \\
\ccmark &  & \ccmark &  &  &  & 72.08 & 63.26 \\
\midrule
 & \ccmark & \ccmark & 4 &  &  & 65.68 & 50.05 \\
 & \ccmark & \ccmark & 4 & \ccmark &  & 72.19 & 59.22 \\
 & \ccmark & \ccmark & 4 &  & \ccmark & 71.56 & 62.60 \\
 & \ccmark & \ccmark & 4 & \ccmark & \ccmark & \textbf{72.03} & \textbf{64.43} \\
 \midrule
 & \ccmark & \ccmark & 9 &  &  & 61.88 & 43.33 \\
 & \ccmark & \ccmark & 9 & \ccmark &  & 67.66 & 52.81 \\
 & \ccmark & \ccmark & 9 &  & \ccmark & 69.58 & 59.01 \\
 & \ccmark & \ccmark & 9 & \ccmark & \ccmark & 70.83 & 60.89 \\
\bottomrule
\vspace{-20pt}
\end{tabular}%
}
\label{tab:ablation_qwen34b}
\end{table*}

\begin{wrapfigure}{r}{0.50\textwidth}
    \centering
    \vspace{-20pt}
    \resizebox{\linewidth}{!}{\includegraphics{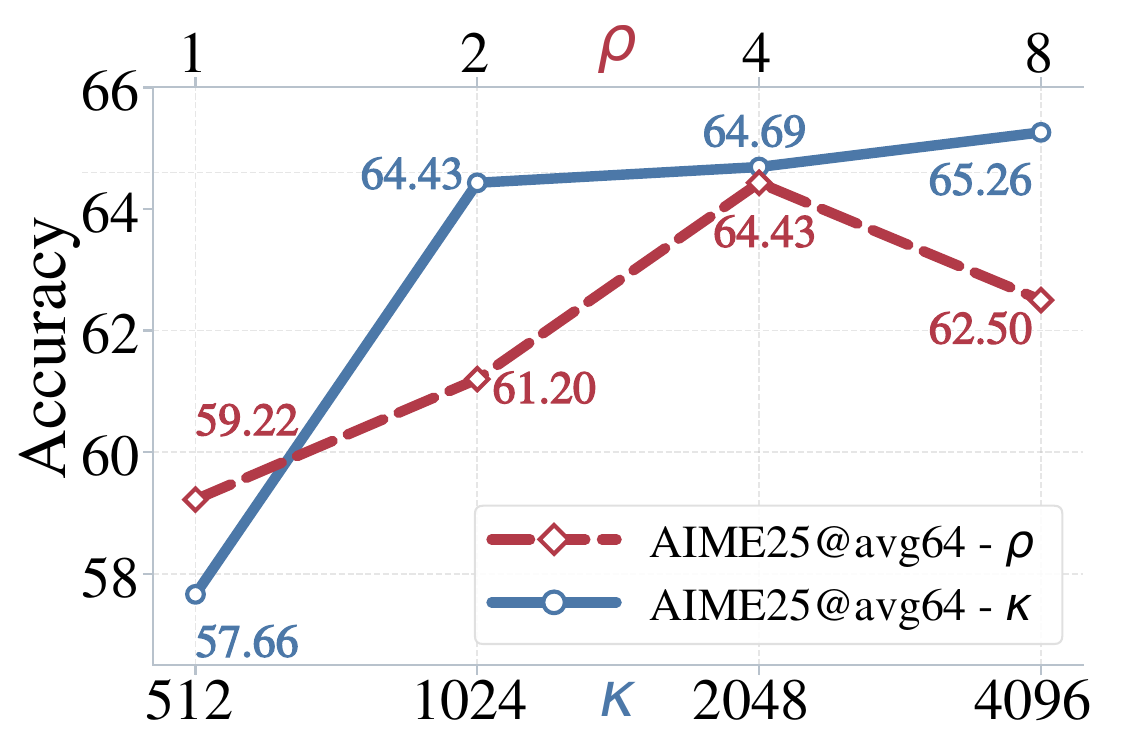}}
    \caption{Sensitivity to proposal expansion factor $\rho$ and sparse budget $\kappa$ on AIME25.}
    \label{fig:ablation_expand_budget}
    \vspace{-8pt}
\end{wrapfigure}

We conduct ablation studies on Qwen3-4B to isolate the effect of each design component. As shown in Table~\ref{tab:ablation_qwen34b}, directly reusing sparse indices across layers leads to a clear accuracy drop, especially on AIME25 ($63.87 \to 50.05$), indicating that rigid cross-layer sharing fails to track layer-wise attention variations. Offline calibration substantially improves the reuse baseline by selecting more suitable proposal/refined layer assignments, while proposal-refined sharing further recovers the lost accuracy by allowing each refined layer to perform local token selection within the inherited candidate region. Combining calibration and proposal-refined sharing gives the best overall result, improving AIME25 from $50.05$ to $64.43$ under the group-4 setting (i.e., $1{:}3$ ratio of proposal to refined layers). The same trend also holds for the more aggressive group-9 setting (i.e., $1{:}8$ ratio of proposal to refined layers), where proposal-refined sharing consistently improves over direct reuse, suggesting that local refinement becomes increasingly important when indices are shared across more layers.

We further study the sensitivity to the proposal expansion factor $\rho$ and sparse budget $\kappa$ in Figure~\ref{fig:ablation_expand_budget}. Increasing $\rho$ from $1$ to $4$ improves accuracy from $59.22$ to $64.43$, confirming that a moderately expanded proposal region is necessary to cover localized cross-layer token shifts. However, further increasing $\rho$ to $8$ slightly degrades performance, suggesting that an overly large proposal may introduce more noisy candidates and weaken the refinement stage. In contrast, increasing the final sparse budget $\kappa$ steadily improves accuracy from $57.66$ at $\kappa=512$ to $65.26$ at $\kappa=4096$, with most of the gain achieved at $\kappa=1024$. These results indicate that $\rho$ between $2$ and $4$, alongside a sparse budget $\kappa$ between $1024$ and $2048$, strikes a highly effective balance between accuracy and efficiency.







\section{Conclusion}

We propose Prism Attention, a proposal-refined sparse attention framework that replaces rigid cross-layer index reuse with a coarse-to-fine strategy. Prism Attention lets proposal layers generate expanded candidate regions and refined layers re-select their own top-$\kappa$ tokens within the corresponding proposed region, at a refinement cost independent of context length; INT2-quantized scoring further reduces the cost of proposal construction. Extensive experiments on reasoning and long-context benchmarks show that Prism achieves the best average accuracy among all evaluated sparse attention methods, while attaining a $3.4\times$ end-to-end decoding speedup over FlashAttention on 128K-token contexts, demonstrating that preserving layer-wise adaptability within a shared proposal region is an effective path toward accurate and efficient long-context inference.

\begin{ack}
We want to thank all the anonymous reviewers for their valuable comments.
This work was supported by the National Natural Science Foundation of China
(NSFC No.~62622608 and 62576232), the National Engineering Research Center of
Deep Learning Technology and Application, and sponsored by CCF-Baidu Open Fund. The authors declare no competing interests.
\end{ack}
{
  \small
  \bibliographystyle{plain}
  \bibliography{main}
}
\clearpage
\appendix
\appendix


\section{Detailed Settings for the Observation Experiments}
We provide the detailed experimental settings of Figure~\ref{fig:forward_layer_offsets} and Figure~\ref{fig:localized_evolution} below. The analysis covers multiple datasets, decoding steps, layers, and attention heads, providing a more comprehensive characterization of the inter-layer
salient-token shift behavior.

\paragraph{Figure~\ref{fig:forward_layer_offsets}.}
\begin{itemize}[leftmargin=*, nosep]
    \item \textbf{Model:} Qwen3-4B with 36 layers and 32 attention heads.
    \item \textbf{Dataset:} AIME24, MATH500, and GPQA; 10 randomly sampled instances per dataset.
    \item \textbf{Context length / decoding traces:} Approximately 1K and 8K
    decoding lengths; 32 consecutive decoding steps per trace (1024--1055 for
    1K and 8192--8223 for 8K).
    \item \textbf{Layer offset:} Reference layers $L_0$--$L_{28}$, with
    offsets 1--7, i.e., comparing $L_l$ and $L_{l+d}$.
    \item \textbf{Averaging procedure:} We compute the overlap ratio for each
    sample, decoding step, attention head, and layer pair, and average over
    the 32 heads, 32 decoding steps, 10 samples, and all valid layer pairs
    with the same offset. Results are reported separately for each dataset.
\end{itemize}

\paragraph{Figure~\ref{fig:localized_evolution}.}
\begin{itemize}[leftmargin=*, nosep]
    \item \textbf{Model:} Qwen3-4B with 36 layers and 32 attention heads.
    \item \textbf{Dataset:} AIME24, MATH500, and GPQA; 10 randomly sampled
    instances per dataset.
    \item \textbf{Context length / decoding traces:} Approximately 8K
    decoding length, with 32 consecutive decoding steps recorded per trace (steps 8192--8223).
    \item \textbf{Averaging procedure:} For each target layer, its top-128
    indices are categorized according to their ranks in the reference layer as
    top-128, 128--511, or outside top-512. The ratios are averaged over the 32
    heads, 32 decoding steps, and all samples. All datasets are pooled
    together for cross-dataset averaging.
\end{itemize}

\section{Generalization of Cross-Layer Attention Locality Across Architectures, Context Lengths, and Task Types}
\label{appd_sec:generalization}

An analysis based only on the Qwen model family is insufficient to establish
whether the observed cross-layer attention characteristics generalize across
model architectures, context lengths, and task types. To address this concern,
we conduct additional experiments along the following three dimensions:

\begin{itemize}[leftmargin=*, nosep]
    \item \textbf{Different model architectures:} Qwen3-4B (GQA),
    DeepSeek-R1-Distill-Llama-8B (GQA), and GLM-4.7-Flash (MLA);
    \item \textbf{Different sequence lengths:} 4K, 8K, and 16K;
    \item \textbf{Different task types:} mathematical reasoning,
    summarization, few-shot learning, and multi-document question answering.
\end{itemize}

Unless otherwise specified, all experiments follow the same analysis setup as
Figure~\ref{fig:localized_evolution}. For each target layer (i.e., Refined
layer), we categorize its own top-128 sparse indices according to their ranks
in the corresponding reference layer (i.e., Proposal layer): \textbf{Prop.
Top-128}, \textbf{Prop. Top-128$\sim$512}, and \textbf{Outside Prop. Top-512}
denote the ratios of target-layer top-128 indices falling in the
reference-layer top-128 set, ranks 128$\sim$511, and outside the top-512 set,
respectively.

\begin{figure}[t]
    \centering
    \begin{minipage}[t]{0.48\linewidth}
        \centering
        \includegraphics[width=\linewidth]{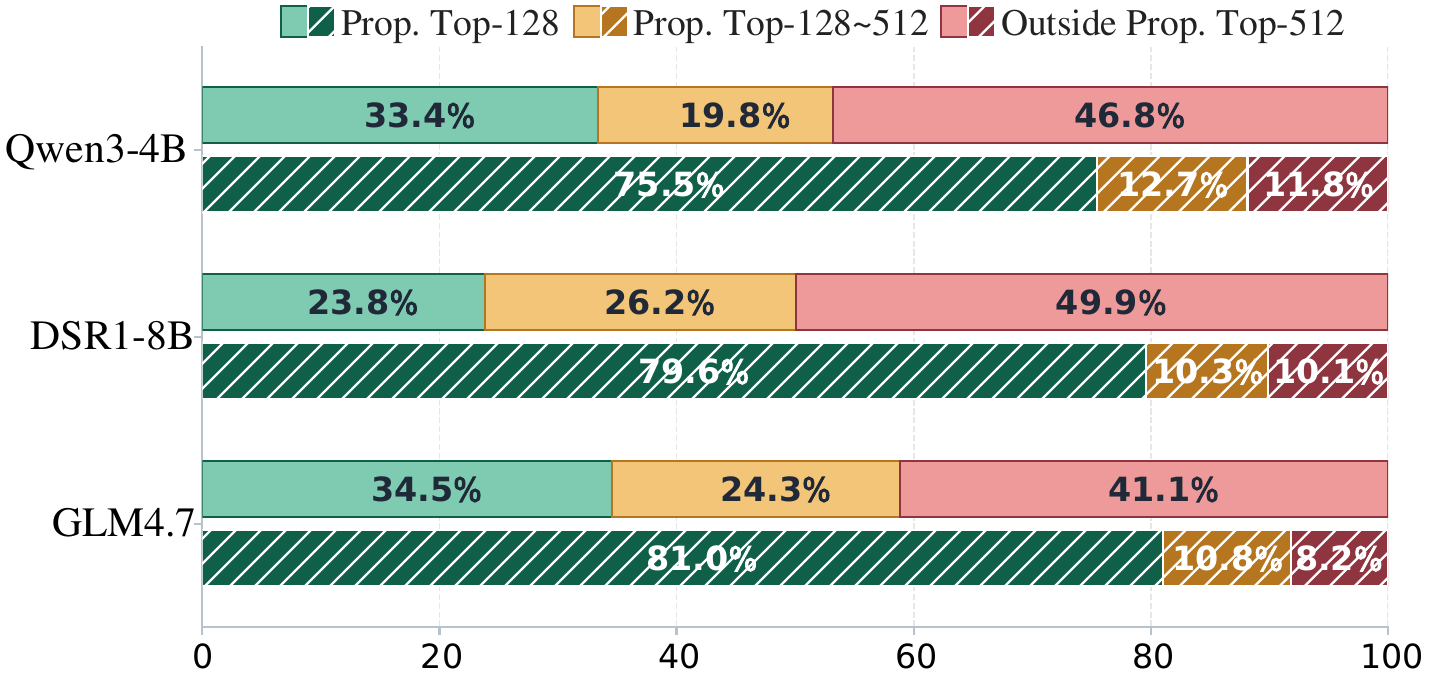}
    \end{minipage}
    \hfill
    \begin{minipage}[t]{0.50\linewidth}
        \centering
        \includegraphics[width=\linewidth]{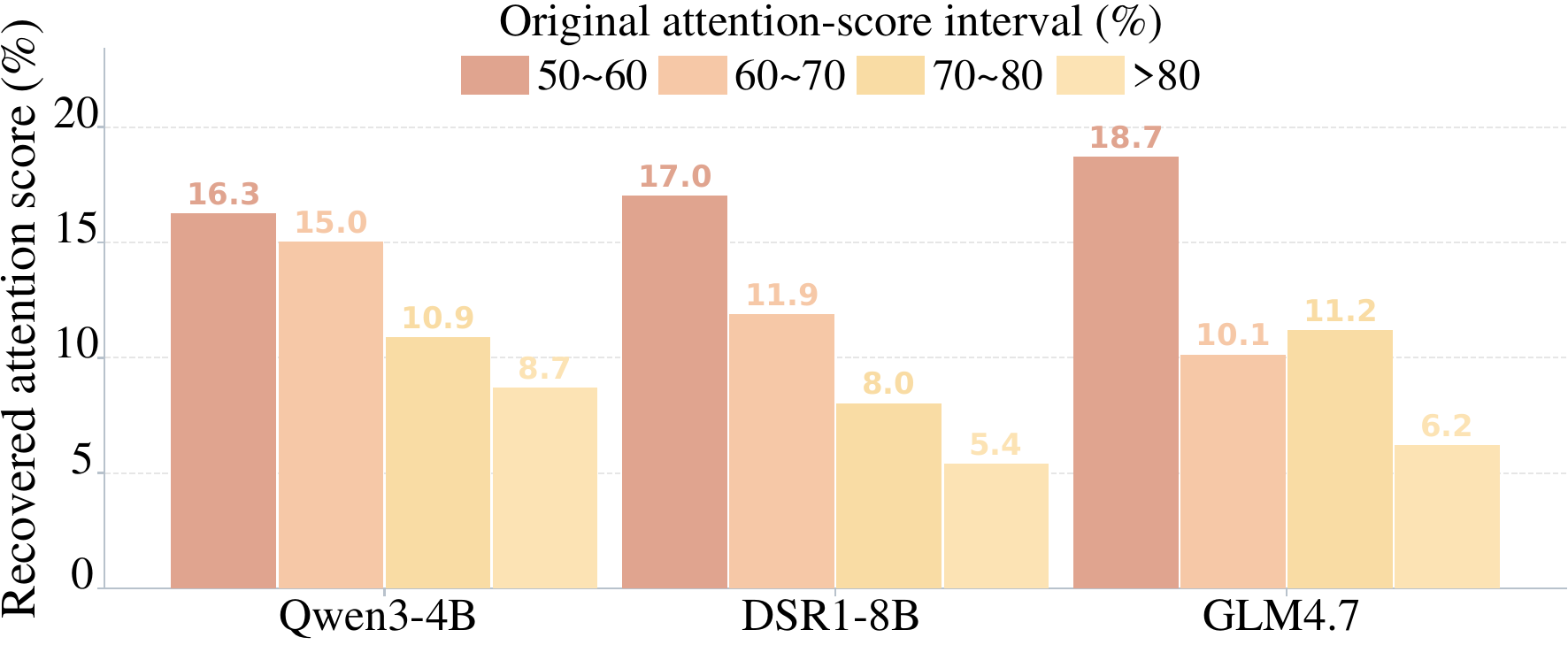}
    \end{minipage}
    \caption{Cross-layer index/attention-mass distribution (Left) and attention score recovered by proposal expansion (Right) across model architectures. \textbf{Left figure}: within each pair of bars, the \textbf{solid} bar denotes the \textbf{index overlap ratio} (the share of the target/refined layer's 128 indices) and the \textbf{hatched} bar denotes the \textbf{attention-mass ratio} (the share of the total attention score those indices carry in the target layer). Both are normalized over the target layer's own Top-128, so the horizontal axis is a percentage and each bar sums to 100\%. \textbf{Right figure}: the legend groups samples by the original attention coverage, i.e., the share of the target layer's Top-128 attention score that is already covered by reusing only the proposal layer's Top-128 indices. The vertical axis reports the additionally recovered attention score (in \%) obtained by expanding the proposal region from Top-128 to Top-512, measured against the same target-layer Top-128 attention total, so both quantities are expressed on a common scale. The recovered attention decreases consistently as the original coverage grows: when the proposal layer's Top-128 already captures most of the target layer's attention, little is left for expansion to recover, whereas in the low-coverage regime---exactly the cases where directly reusing the proposal indices would be most lossy---the expanded region recovers the largest amount of attention score. \textbf{Note:} for conciseness, we abbreviate DeepSeek-R1-Distill-Llama-8B as DSR1-8B and GLM-4.7-Flash as GLM4.7 in both figures above.}
    \label{fig:models}
    \vspace{-10pt}
\end{figure}

\begin{figure}[t]
    \centering
    \begin{minipage}[t]{0.48\linewidth}
        \centering
        \includegraphics[width=\linewidth]{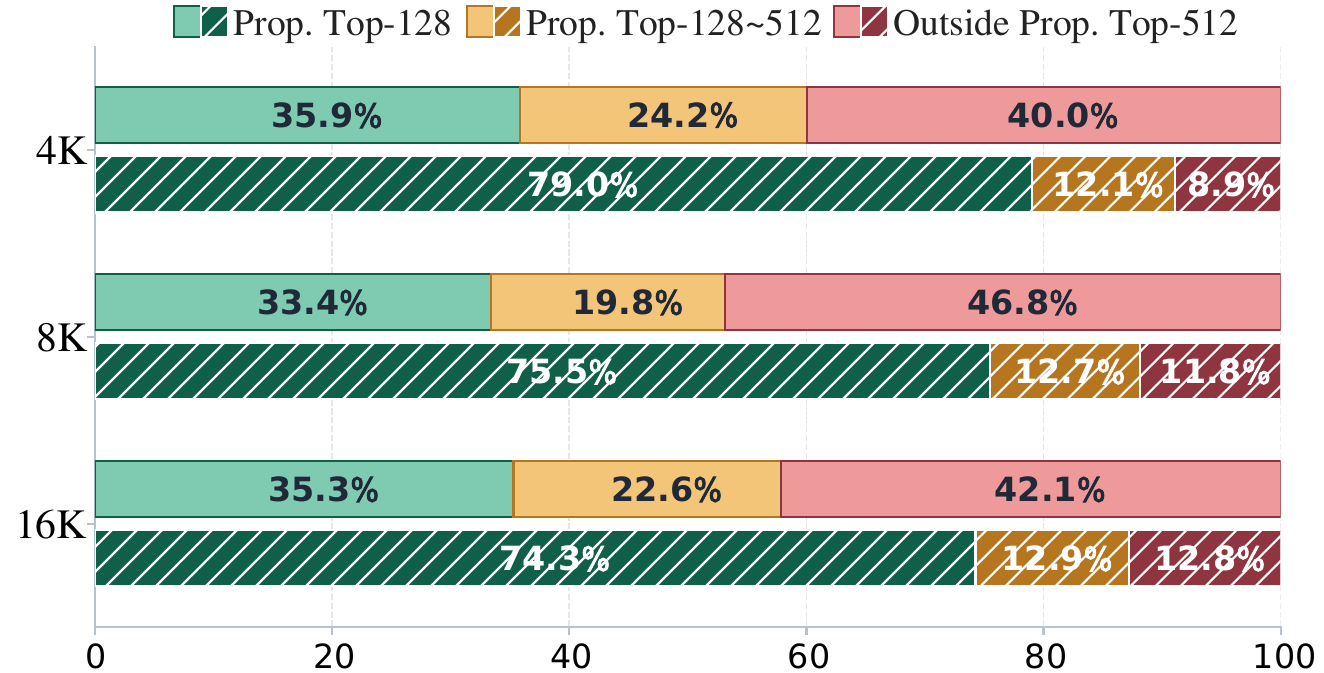}
    \end{minipage}
    \hfill
    \begin{minipage}[t]{0.50\linewidth}
        \centering
        \includegraphics[width=\linewidth]{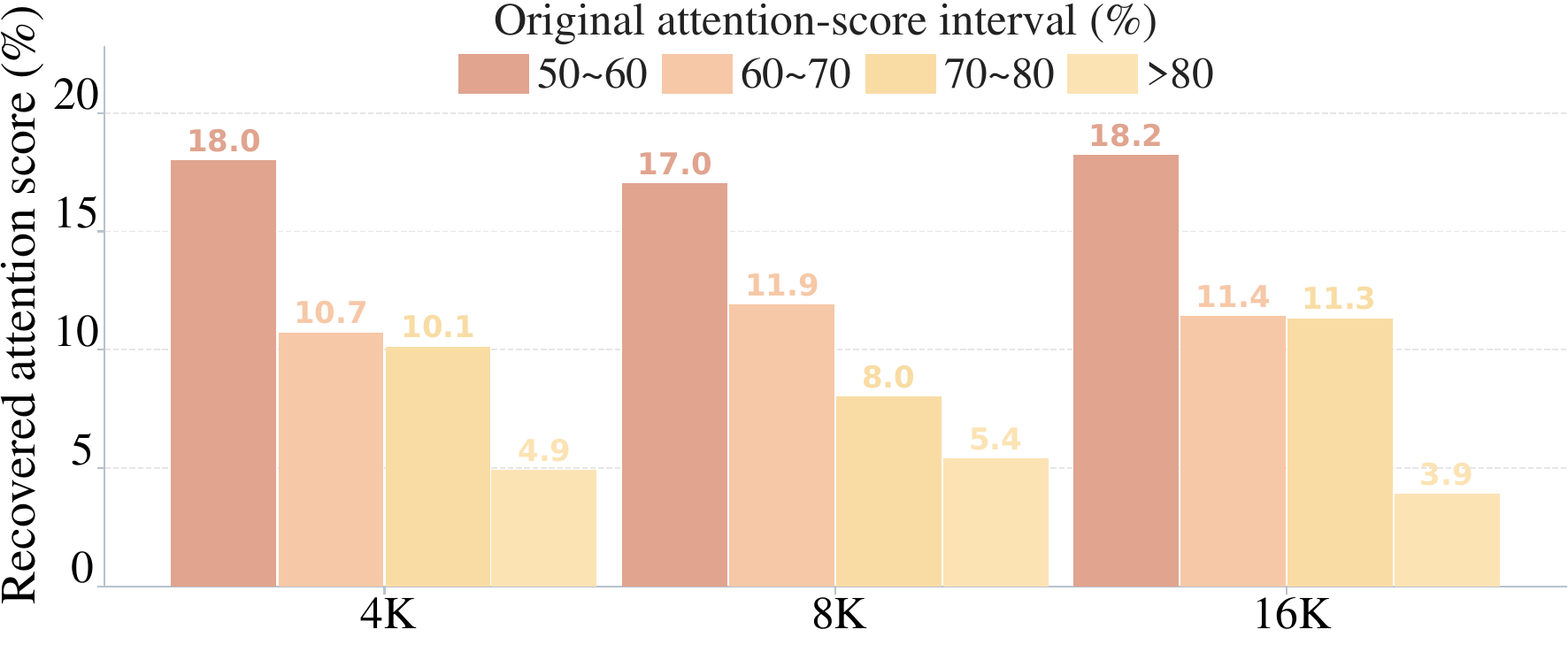}
    \end{minipage}
    \caption{Cross-layer index/attention-mass distribution (Left) and attention score recovered by proposal expansion (Right) across context lengths. Both follow the conventions of Figure~\ref{fig:models}.}
    \label{fig:lengths}
    \vspace{-8pt}
\end{figure}

\begin{figure}[t]
    \centering
    \begin{minipage}[t]{0.48\linewidth}
        \centering
        \includegraphics[width=\linewidth]{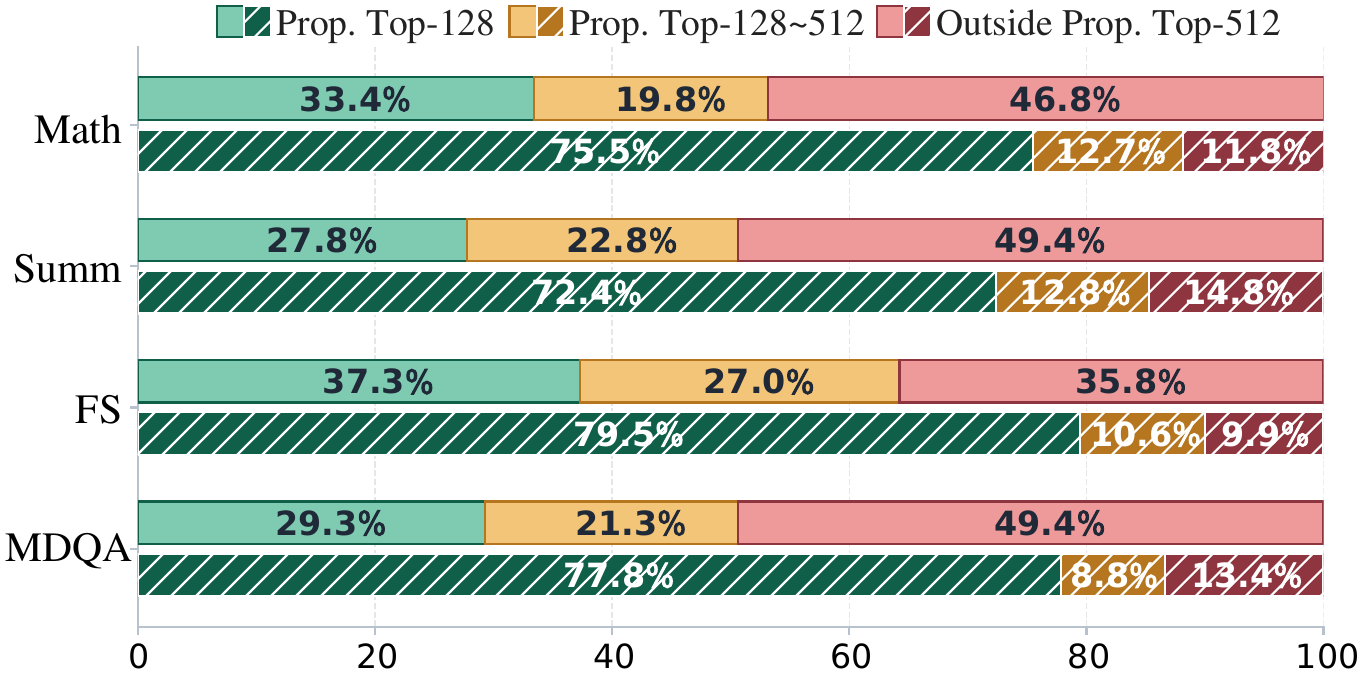}
    \end{minipage}
    \hfill
    \begin{minipage}[t]{0.50\linewidth}
        \centering
        \includegraphics[width=\linewidth]{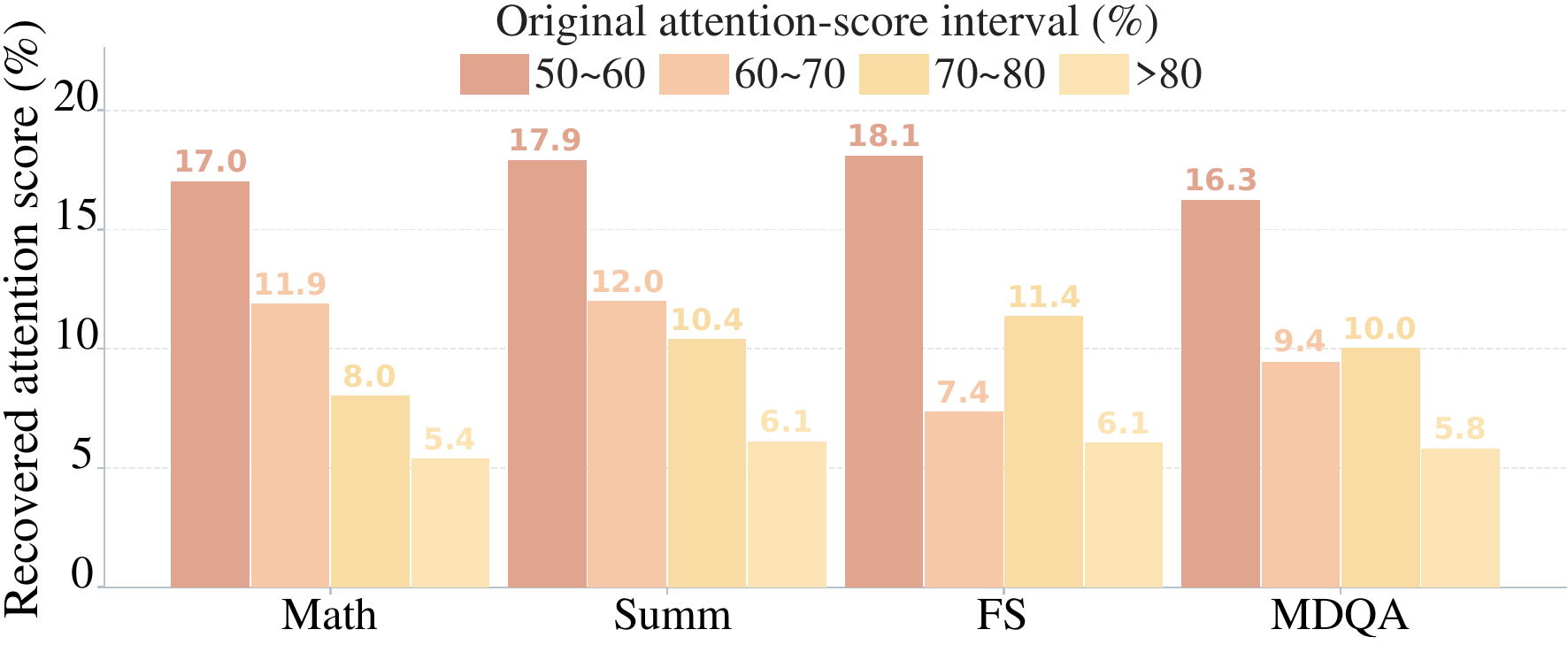}
    \end{minipage}
    \caption{Cross-layer index/attention-mass distribution (Left) and attention score recovered by proposal expansion (Right) across task types. Both follow the conventions of Figure~\ref{fig:models}. \textbf{Note:} for conciseness, we abbreviate Mathematical Reasoning, Summarization, Few-shot Learning, and Multi-Document QA as Math, Summ, FS, and MDQA, respectively, in both figures above.}
    \label{fig:tasks}
    \vspace{-8pt}
\end{figure}

\subsection{Different Model Architectures}

To examine whether this phenomenon generalizes beyond the Qwen model family, we additionally evaluate DeepSeek-R1-Distill-Llama-8B and GLM-4.7-Flash on 8K-token mathematical reasoning sequences, with Qwen3-4B included as a reference. Figure~\ref{fig:models} demonstrates that the localized cross-layer drift observed in our study not only occurs in Qwen but also across different model architectures and attention mechanisms, including both GQA and MLA, and that proposal expansion continues to recover additional attention scores under the same trend across all evaluated architectures.

\subsection{Different Sequence Lengths}

To investigate whether the locality of cross-layer attention shifts changes with context length, we further conduct experiments on Qwen3-4B using mathematical reasoning sequences of 4K, 8K, and 16K tokens. All other analysis settings are kept unchanged. Figure~\ref{fig:lengths} indicates that both the localized cross-layer shift pattern and the benefit of proposal expansion in recovering additional attention score hold robustly across all evaluated context lengths.

\begin{figure*}
    \centering
    \begin{subfigure}[t]{0.47\textwidth}
        \centering
        \resizebox{\textwidth}{!}{\includegraphics{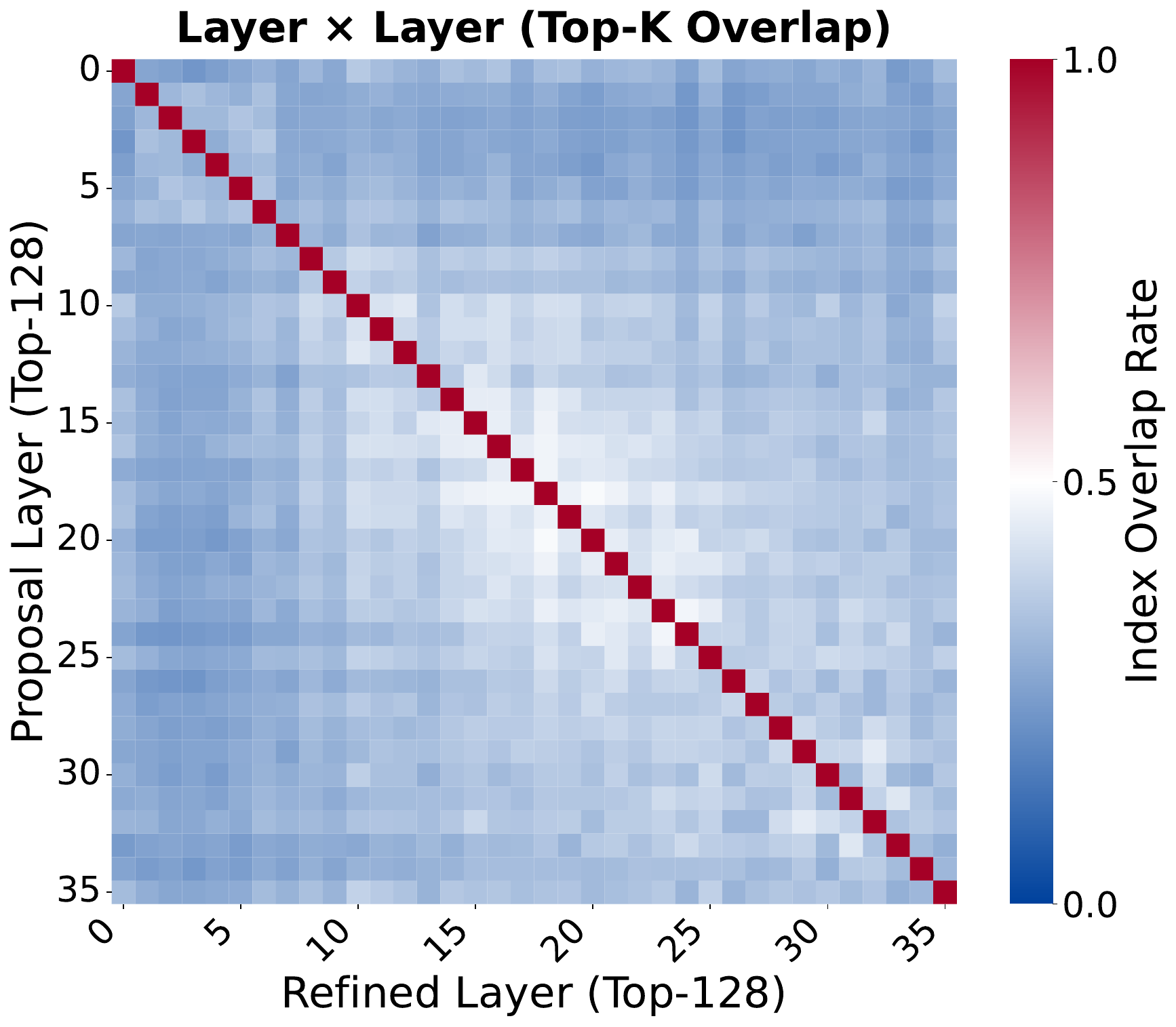}}
        \caption{GPQA, proposal$=$top-128.}
        \label{fig:overlap_gpqa_128}
    \end{subfigure}
    \hspace{0.5cm}
    \begin{subfigure}[t]{0.47\textwidth}
        \centering
        \resizebox{\textwidth}{!}{\includegraphics{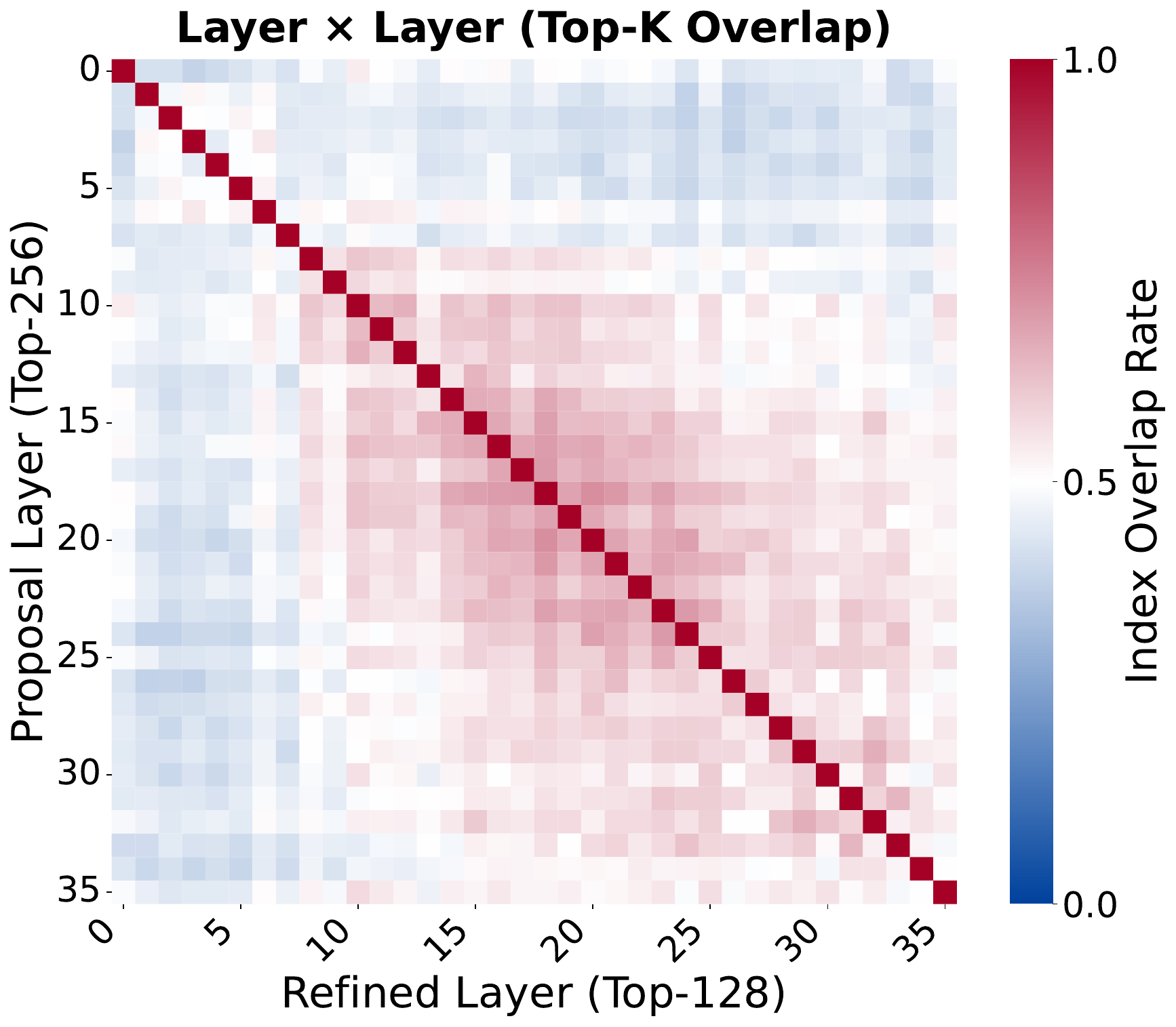}}
        \caption{GPQA, proposal$=$top-256.}
        \label{fig:overlap_gpqa_256}
    \end{subfigure}

    \vspace{2pt}   
    \begin{subfigure}[t]{0.47\textwidth}
        \centering
        \resizebox{\textwidth}{!}{\includegraphics{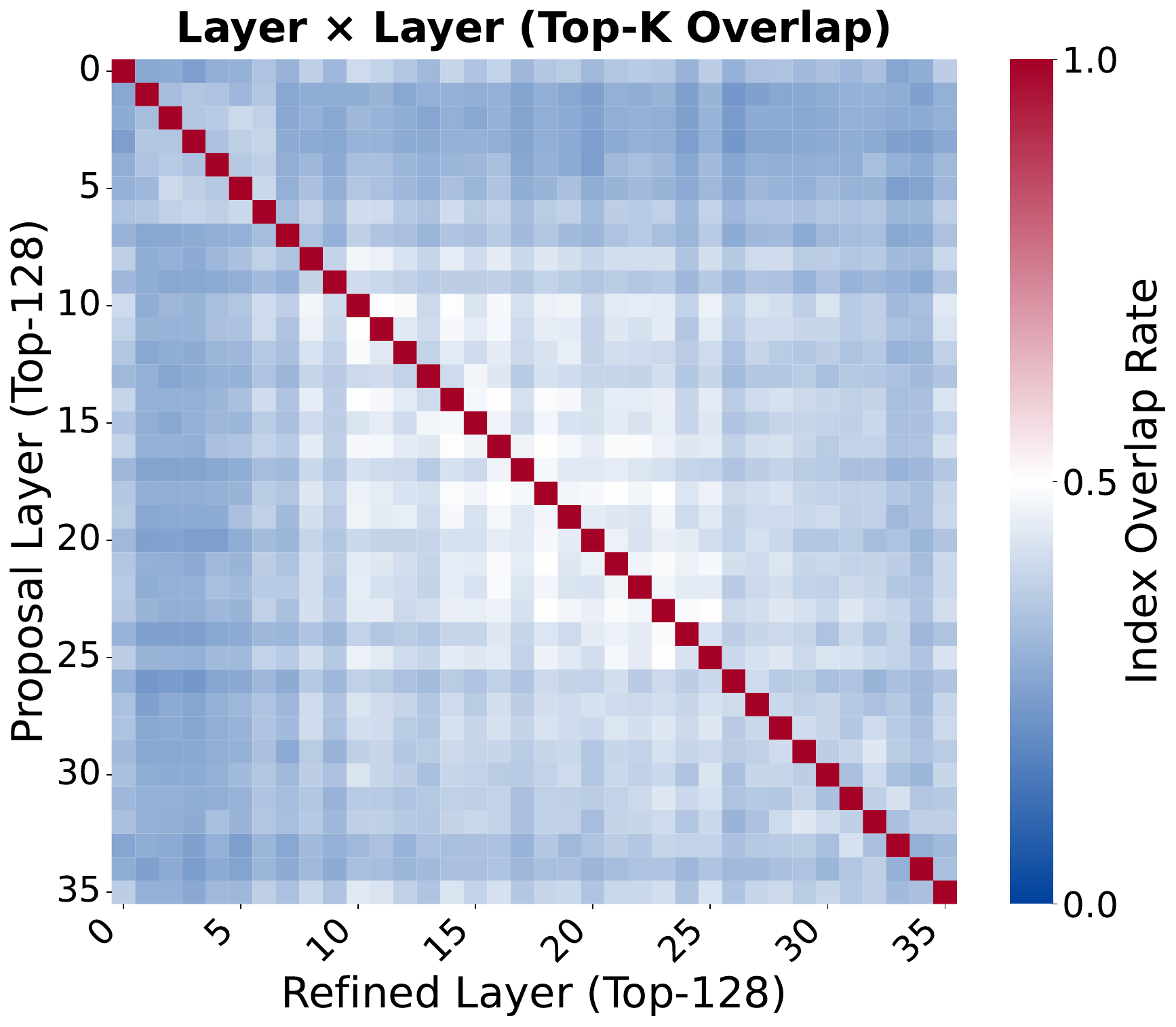}}
        \caption{MATH500, proposal$=$top-128.}
        \label{fig:overlap_math_128}
    \end{subfigure}
    \hspace{0.5cm}
    \begin{subfigure}[t]{0.47\textwidth}
        \centering
        \resizebox{\textwidth}{!}{\includegraphics{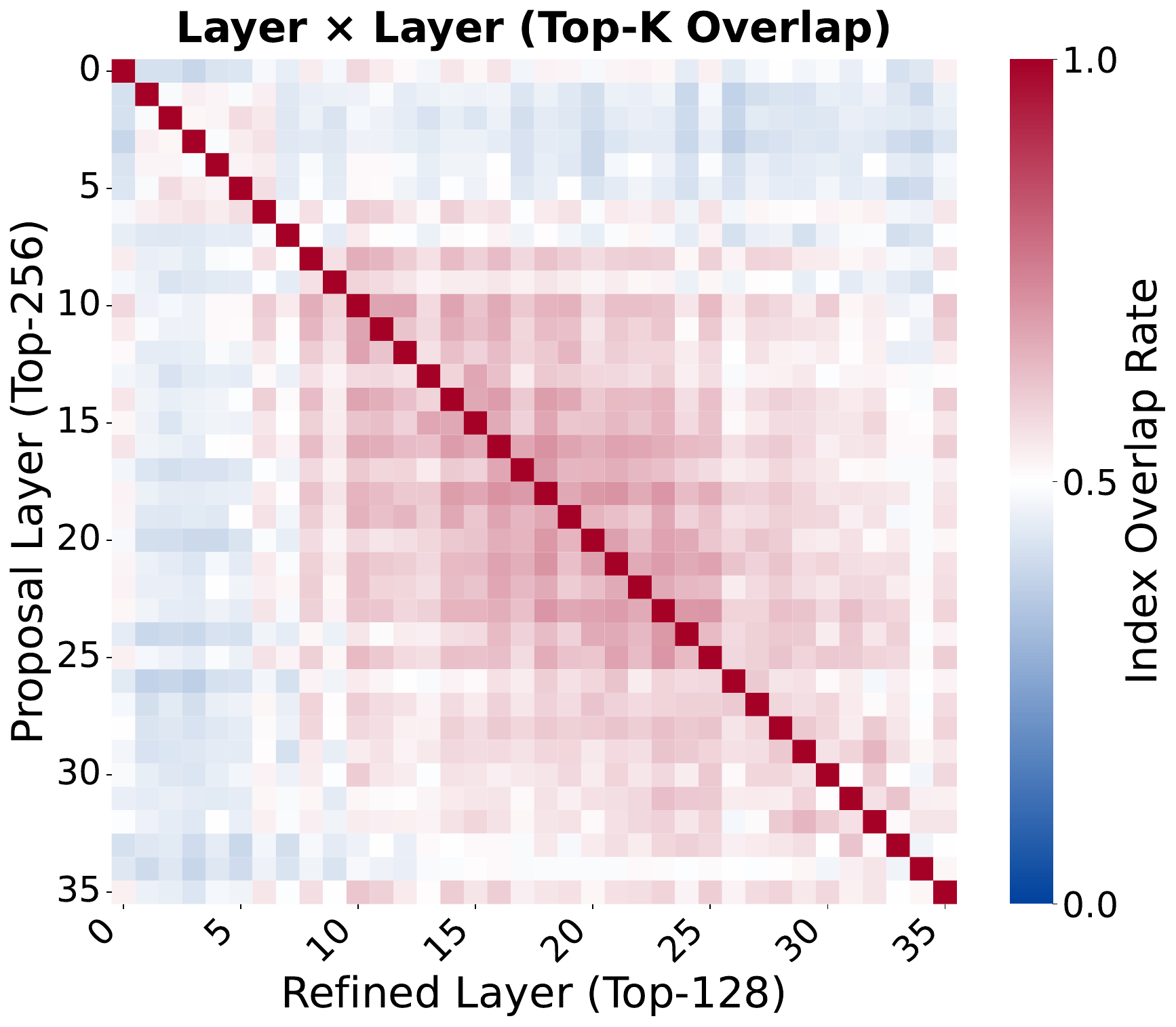}}
        \caption{MATH500, proposal$=$top-256.}
        \label{fig:overlap_math_256}
    \end{subfigure}

    \caption{
    {Effect of expanded proposal regions on cross-layer sparse-index coverage, with the refined-layer sparse support fixed at top-128 (Qwen3-4B; GPQA~\cite{gpqa} and MATH500~\cite{math500} traces).}
    }
    \label{fig:index_overlap_heatmap_add}
    \vspace{-10pt}
\end{figure*}

\subsection{Different Task Types}

To verify that this observation is not specific to mathematical reasoning, we further analyze four contextual task categories using Qwen3-4B, with all tasks based on 8K-token sequences:

\begin{itemize}[leftmargin=*, nosep]
    \item \textbf{Mathematical Reasoning:} 10 samples are randomly selected
    from each of AIME24, MATH500, and GPQA;
    \item \textbf{Summarization:} 10 samples are randomly selected from the
    summarization task category in LongBench;
    \item \textbf{Few-shot Learning:} 10 samples are randomly selected from
    the few-shot learning task category in LongBench;
    \item \textbf{Multi-Document QA:} 10 samples are randomly selected from
    the multi-document question-answering task category in LongBench.
\end{itemize}

As shown in Figure~\ref{fig:tasks}, different tasks exhibit different degrees of cross-layer index variation, indicating that task type affects cross-layer attention dynamics. Nevertheless, despite these differences in the magnitude of cross-layer drift, the expanded proposal region consistently covers a set of candidate tokens with substantial attention contributions across all evaluated tasks. These results demonstrate that the localized cross-layer attention shifts observed in our study are not limited to mathematical reasoning, but also occur in summarization, few-shot learning, and multi-document question answering, which exhibit substantially different context-access patterns.

To further verify the universality of our observations on AIME24, we present additional index-overlap visualizations for the GPQA and MATH500 datasets. As shown in Figure~\ref{fig:index_overlap_heatmap_add}, attention index dynamics are highly consistent regardless of the task domain. Rigid index reuse (left panels, top-$128$) yields poor overlap for distant layers, reflecting how salient tokens inherently shift across network depth. Conversely, expanding the proposal pool to top-$256$ ($\rho=2$, right panels) effectively preserves cross-layer coverage. This confirms that inter-layer token shifting is a universal phenomenon addressed by our localized expansion strategy across various applications.

\section{Sensitivity to Calibration Sample Selection}

\begin{figure}
    \centering
    \includegraphics[width=\linewidth]{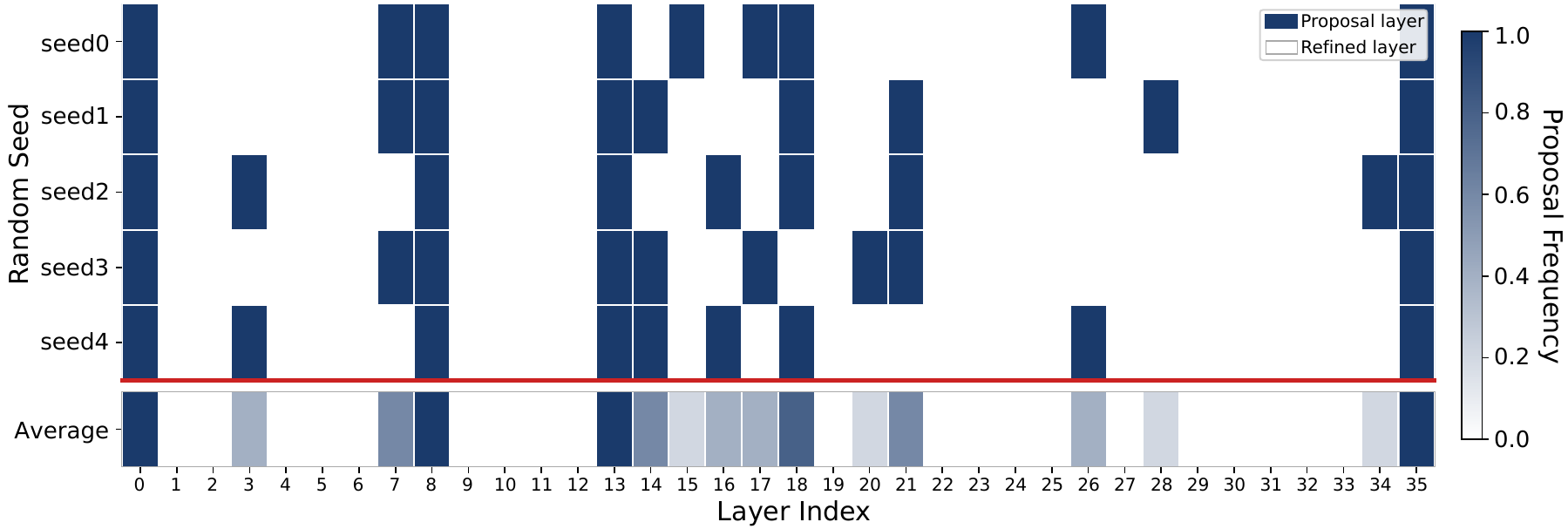} 
    \caption{Sensitivity of proposal layer assignment to calibration sample selection on Qwen3-4B.} 
    \label{fig:proposal_layers_heatmap}
\end{figure}

We perform sensitivity analyses regarding the selection of calibration samples. As illustrated in Figure~\ref{fig:proposal_layers_heatmap}, the majority of proposal layers exhibit high stability across different sample subsets selected via various random seeds. Notably, layers 0, 8, 13, and 35 are almost always designated as proposal layers, which is broadly consistent with the searched patterns reported in Table~\ref{tab:searched_patterns}. These observations underscore the robustness and consistency of the offline calibration procedure.

To further validate these findings, we evaluate the downstream performance of the varying layer assignment patterns generated across different random seeds. We maintain a sparse budget of $\kappa=1024$ and an expansion factor of $\rho=4$. The results, presented in Table~\ref{tab:seed_sensitivity}, indicate that the final accuracies achieved by these different patterns remain remarkably stable. This consistent downstream performance further demonstrates that our proposal-refined mechanism is inherently robust to minor fluctuations in layer assignments, and that the offline calibration reliably captures generalized attention redundancy without overfitting to the specific calibration subsets.

\begin{table}[t]
\centering
\caption{Searched Proposal/Refined layer assignment patterns
(\textbf{F}: Proposal layer, \textbf{S}: Refined layer).}
\label{tab:searched_patterns}
\renewcommand{\arraystretch}{1.3}
\begin{tabular}{llc}
\toprule
{Model} & {Config} & {Pattern} \\
\midrule
\multirow{2}{*}{Qwen3-4B} & group-4 & $\mathcal{FSSSSSSFFSSSSFSFSFFSSSSSSSFSSSSSSSSF}$ \\
                           & group-9 & $\mathcal{FSSSSSSFSSSSSFSSSSFSSSSSSSSSSSSSSSSS}$ \\
\midrule
\multirow{2}{*}{Qwen3-8B} & group-4 & $\mathcal{FSSSSSSFFSSSSFFSSSFSFFSSSSSSFSSSSSSS}$ \\
                           & group-9 & $\mathcal{FSSSSSSFSSSSSFFSSSSSSSSSSSSSSSSSSSSS}$ \\
\bottomrule
\end{tabular}
\end{table}

\begin{table}[t]
\centering
\caption{Performance across layer assignment patterns from random calibration seeds.}
\label{tab:seed_sensitivity}
\begin{tabular}{lccccc}
\toprule
{Metric} & {Seed 0} & {Seed 1} & {Seed 2} & {Seed 3} & {Seed 4} \\
\midrule
AIME25      & 64.43  & 64.32  & 64.69  & 64.06  & 63.91  \\
\bottomrule
\end{tabular}
\end{table}

\section{Complete Efficiency Results: From Kernels to End-to-End}

\begin{figure*}
    \centering
    \begin{subfigure}[t]{0.48\textwidth}  
        \centering
        \resizebox{\textwidth}{!}{\includegraphics{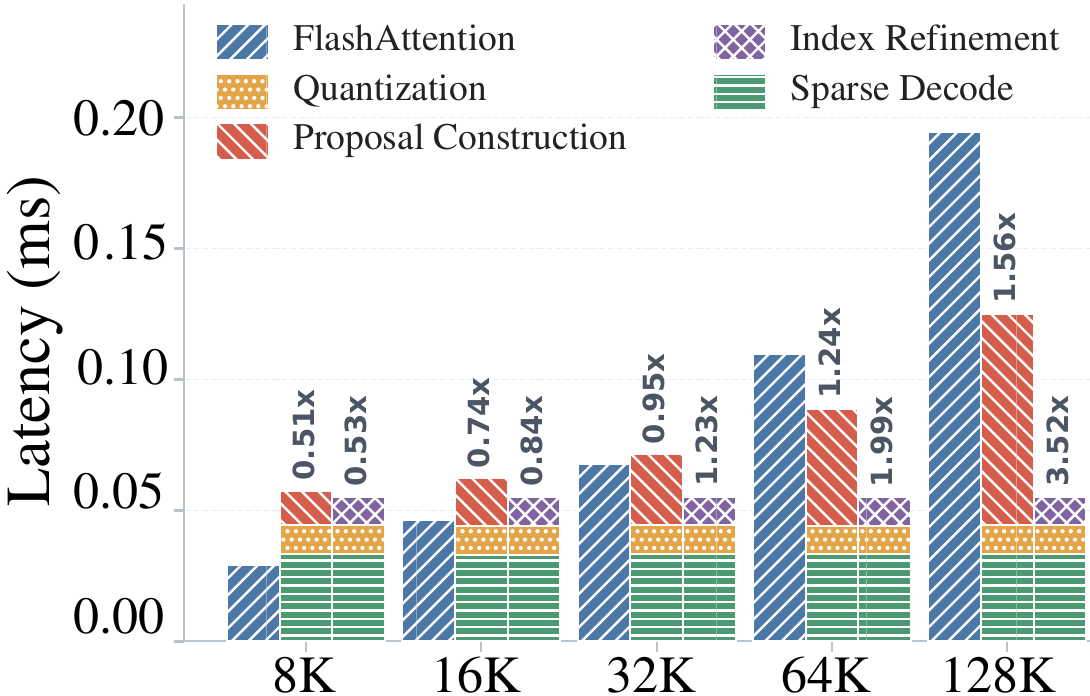}}  
        \caption{Batch Size = 1}
        \label{fig:model_lantency_a}
    \end{subfigure}
    \hfill  
    \begin{subfigure}[t]{0.48\textwidth}
        \centering
        \resizebox{\textwidth}{!}{\includegraphics{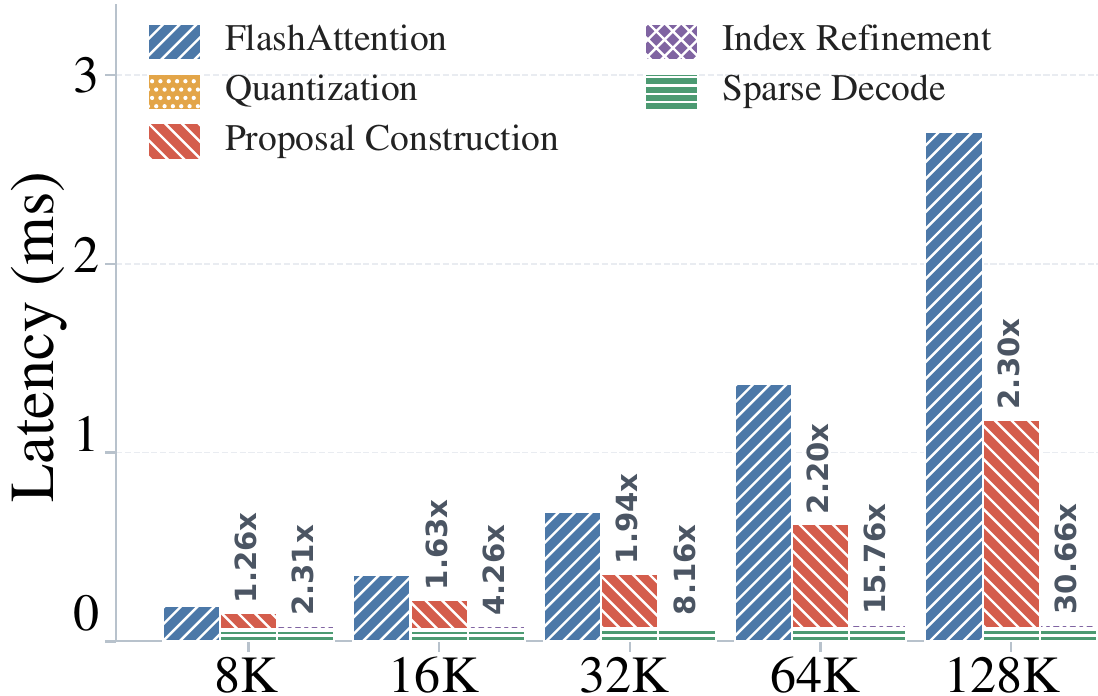}}  
        \caption{Batch Size = 16}
        \label{fig:model_lantency_b}
    \end{subfigure}
\caption{Component-wise latency breakdown of Prism Attention compared to the FlashAttention baseline across different sequence lengths at (a) BSZ=1 and (b) BSZ=16.}
\label{fig:model_lantency}
\end{figure*}

\textbf{Latency of individual kernel components.}
We first measure our customized Triton kernels in isolation, timing each attention-path component separately under CUDA Graph replay at the full head configuration of Qwen3-4B, without tensor-parallel sharding and without any model-level or framework-level overhead. Figure~\ref{fig:model_lantency} provides additional measurements for BSZ=1 and BSZ=16, and Table~\ref{tab:detailed_latency} reports the exact execution time of every component alongside the FlashAttention baseline. The decomposition shows that index refinement stays essentially constant at roughly $0.011$ ms from 8K to 128K, in sharp contrast to proposal construction, which scales with context length, confirming that the local refinement step performed by every refined layer incurs negligible and length-independent cost, and that the length-sensitive cost is confined to proposal layers.

\textbf{Per-layer latency under the SGLang integration.}
We further evaluate the same kernels after integrating them into SGLang, where each layer is timed as a single fused CUDA Graph chain rather than as separate components, using the tensor shapes of one TP8 rank ($4$ query heads, $1$ KV head, head dimension $128$) and a $2$K sparse budget. Table~\ref{tab:kernel-latency-breakdown_sglang} reports the cost of an index-construction layer (Gate) and an index-reuse layer (Refine), together with the effective per-layer cost under the $1{:}3$ ratio adopted throughout our experiments. The reuse-layer cost again remains nearly flat across all context lengths while the construction-layer cost grows only mildly, so the effective cost scales far more slowly than dense FlashAttention and reaches a $5.42\times$ speedup at 128K with BSZ$=16$. Prism carries a modest overhead relative to IndexCache at short contexts and small batches, since INT2 quantization and the enlarged candidate pool introduce a fixed cost that is not yet amortized; this gap narrows as the workload grows and reverses at 128K with BSZ$=16$, where Prism's cheaper index construction makes it the faster of the two.

\textbf{End-to-end serving latency.} Table~\ref{tab:e2e-latency} provides complete decoding latency measured inside the SGLang framework, so that the kernel-level gains above can be assessed under a realistic deployment.


\begin{table*}[t]
\centering
\caption{Detailed latency breakdown (ms) of Prism Attention kernels compared with FlashAttention across various batch sizes and sequence lengths.}
\label{tab:detailed_latency}
\small
\begin{tabular}{llccccc}
\toprule
\textbf{Batch Size} & \textbf{Component} & \textbf{8K} & \textbf{16K} & \textbf{32K} & \textbf{64K} & \textbf{128K} \\
\midrule
\multirow{5}{*}{BSZ=1} & FlashAttention & 0.0293 & 0.0463 & 0.0679 & 0.1097 & 0.1945 \\
\cmidrule{2-7}
 & Quantization & 0.0111 & 0.0109 & 0.0111 & 0.0109 & 0.0109 \\
 & Proposal Construction & 0.0132 & 0.0182 & 0.0270 & 0.0446 & 0.0807 \\
 & Index Refinement & 0.0108 & 0.0111 & 0.0109 & 0.0109 & 0.0109 \\
 & Sparse Decode & 0.0333 & 0.0331 & 0.0333 & 0.0332 & 0.0334 \\
\midrule
\multirow{5}{*}{BSZ=4} & FlashAttention & 0.0644 & 0.1066 & 0.1911 & 0.3617 & 0.6985 \\
\cmidrule{2-7}
 & Quantization & 0.0109 & 0.0110 & 0.0110 & 0.0110 & 0.0109 \\
 & Proposal Construction & 0.0268 & 0.0459 & 0.0831 & 0.1503 & 0.2930 \\
 & Index Refinement & 0.0109 & 0.0108 & 0.0109 & 0.0109 & 0.0106 \\
 & Sparse Decode & 0.0340 & 0.0337 & 0.0337 & 0.0335 & 0.0340 \\
\midrule
\multirow{5}{*}{BSZ=16} & FlashAttention & 0.1864 & 0.3539 & 0.6871 & 1.3633 & 2.6978 \\
\cmidrule{2-7}
 & Quantization & 0.0110 & 0.0109 & 0.0111 & 0.0110 & 0.0107 \\
 & Proposal Construction & 0.0824 & 0.1492 & 0.2853 & 0.5482 & 1.1019 \\
 & Index Refinement & 0.0149 & 0.0152 & 0.0151 & 0.0150 & 0.0150 \\
 & Sparse Decode & 0.0547 & 0.0569 & 0.0580 & 0.0605 & 0.0623 \\
\bottomrule
\end{tabular}
\end{table*}

\begin{table}[t]
\centering
\footnotesize
\setlength{\tabcolsep}{3.5pt}
\renewcommand{\arraystretch}{1.1}
\caption{Attention-path kernel latency breakdown under a $2$K sparse
budget, measured with our Triton kernels integrated into the
\textbf{SGLang} serving framework. 
\textbf{Gate} and \textbf{Refine} denote the full layer chain of an
index-construction layer (Prism: proposal, IndexCache: anchor) and an
index-reuse layer (Prism: refined, IndexCache: shared), respectively.
\textbf{Eff.} is the effective cost at the $1{:}3$
construction-to-reuse ratio used throughout our experiments, i.e.,\
$T=(T_{\mathrm{construct}}+3\,T_{\mathrm{reuse}})/4$; parenthesized
values are speedups over FlashAttention (FA). All latencies in ms.}
\label{tab:kernel-latency-breakdown_sglang}
\begin{tabular}{ll rrr rrr r}
\toprule
& & \multicolumn{3}{c}{Prism} & \multicolumn{3}{c}{IndexCache} & \\
\cmidrule(lr){3-5}\cmidrule(lr){6-8}
BSZ & Seq & Gate & Refine & Eff. & Gate & Reuse & Eff. & FA \\
\midrule
\multirow{5}{*}{$1$}
& 8K   & 0.0328 & 0.0208 & 0.0238 (0.61$\times$) & 0.0279 & 0.0116 & 0.0157 (0.93$\times$) & 0.0145 \\
& 16K  & 0.0491 & 0.0198 & 0.0271 (0.67$\times$) & 0.0561 & 0.0116 & 0.0227 (0.80$\times$) & 0.0182 \\
& 32K  & 0.0490 & 0.0196 & 0.0269 (0.79$\times$) & 0.0565 & 0.0117 & 0.0229 (0.93$\times$) & 0.0213 \\
& 64K  & 0.0518 & 0.0197 & 0.0278 (0.89$\times$) & 0.0578 & 0.0115 & 0.0231 (1.07$\times$) & 0.0247 \\
& 128K & 0.0545 & 0.0198 & 0.0285 (1.50$\times$) & 0.0603 & 0.0118 & 0.0240 (1.78$\times$) & 0.0426 \\
\midrule
\multirow{5}{*}{$8$}
& 8K   & 0.0377 & 0.0233 & 0.0269 (0.81$\times$) & 0.0321 & 0.0123 & 0.0173 (1.26$\times$) & 0.0218 \\
& 16K  & 0.0666 & 0.0222 & 0.0333 (1.17$\times$) & 0.0553 & 0.0124 & 0.0231 (1.68$\times$) & 0.0388 \\
& 32K  & 0.0653 & 0.0221 & 0.0329 (1.81$\times$) & 0.0711 & 0.0123 & 0.0270 (2.20$\times$) & 0.0594 \\
& 64K  & 0.0846 & 0.0223 & 0.0379 (2.70$\times$) & 0.0944 & 0.0121 & 0.0327 (3.14$\times$) & 0.1025 \\
& 128K & 0.1169 & 0.0226 & 0.0461 (4.10$\times$) & 0.1393 & 0.0123 & 0.0441 (4.29$\times$) & 0.1892 \\
\midrule
\multirow{5}{*}{$16$}
& 8K   & 0.0432 & 0.0270 & 0.0311 (1.22$\times$) & 0.0417 & 0.0133 & 0.0204 (1.86$\times$) & 0.0379 \\
& 16K  & 0.0731 & 0.0262 & 0.0380 (1.55$\times$) & 0.0747 & 0.0134 & 0.0287 (2.05$\times$) & 0.0589 \\
& 32K  & 0.0905 & 0.0268 & 0.0427 (2.38$\times$) & 0.0979 & 0.0135 & 0.0346 (2.94$\times$) & 0.1017 \\
& 64K  & 0.1230 & 0.0281 & 0.0518 (3.60$\times$) & 0.1432 & 0.0135 & 0.0459 (4.06$\times$) & 0.1863 \\
& 128K & 0.1794 & 0.0279 & 0.0658 (5.42$\times$) & 0.2326 & 0.0137 & 0.0684 (5.22$\times$) & 0.3569 \\
\bottomrule
\end{tabular}
\end{table}

\begin{table}[t]
\centering
\small
\setlength{\tabcolsep}{4pt}
\renewcommand{\arraystretch}{1.15}
\caption{End-to-end per-step decoding latency on Qwen3-4B with the
SGLang serving framework under a $2$K sparse budget. Each cell reports
the absolute latency (ms) and, in parentheses, the speedup over
FlashAttention (FA).}
\label{tab:e2e-latency}
\resizebox{\linewidth}{!}{%
\begin{tabular}{l cc cc cc}
\toprule
& \multicolumn{2}{c}{Seq $=16$K}
& \multicolumn{2}{c}{Seq $=64$K}
& \multicolumn{2}{c}{Seq $=128$K} \\
\cmidrule(lr){2-3}\cmidrule(lr){4-5}\cmidrule(lr){6-7}
Method
& BSZ=1 & BSZ=16
& BSZ=1 & BSZ=16
& BSZ=1 & BSZ=16 \\
\midrule
FA      & 2.151 (1.00$\times$) & 4.057 (1.00$\times$) & 2.623 (1.00$\times$) & 8.748 (1.00$\times$) & 3.074 (1.00$\times$) & 15.025 (1.00$\times$) \\
Seer-R  & 2.520 (0.85$\times$) & 3.101 (1.31$\times$) & 2.580 (1.02$\times$) & 3.208 (2.73$\times$) & 2.659 (1.16$\times$) & 3.483 (4.31$\times$) \\
LIM     & 2.399 (0.90$\times$) & 3.465 (1.17$\times$) & 2.517 (1.04$\times$) & 4.166 (2.10$\times$) & 2.667 (1.15$\times$) & 5.147 (2.92$\times$) \\
IDC-g4  & 2.261 (0.95$\times$) & 2.854 (1.42$\times$) & 2.317 (1.13$\times$) & 3.550 (2.46$\times$) & 2.450 (1.26$\times$) & 4.434 (3.39$\times$) \\
IDC-g9  & 2.026 (1.06$\times$) & 2.557 (1.59$\times$) & 2.062 (1.27$\times$) & 2.912 (3.00$\times$) & 2.156 (1.43$\times$) & 3.348 (4.49$\times$) \\
Prism-g4 & 2.492 (0.86$\times$) & 3.302 (1.23$\times$) & 2.565 (1.02$\times$) & 3.835 (2.28$\times$) & 2.678 (1.15$\times$) & 4.467 (3.36$\times$) \\
Prism-g9 & 2.314 (0.93$\times$) & 3.091 (1.31$\times$) & 2.374 (1.11$\times$) & 3.343 (2.62$\times$) & 2.474 (1.24$\times$) & 3.703 (4.06$\times$) \\
\bottomrule
\end{tabular}}
\end{table}

\begin{algorithm}[t]
\caption{PrismAttention Decoding at Step $t$}
\label{alg:prism}
\begin{algorithmic}[1]
\Require Input representation $X_t$; layers $1,\dots,N$ partitioned into
proposal layers $\mathcal{F}$ and refined layers $\mathcal{S}$; association
function $f:\mathcal{S}\!\to\!\mathcal{F}$; sparse budget $\kappa$; expansion
factor $\rho\ge 1$.
\Statex \textit{// $\mathcal{C}_t^{\mathcal{F}}$: shared candidate cache populated by proposal layers.}
\For{$\ell = 1$ \textbf{to} $N$}
  \If{$\ell \in \mathcal{F}$} \Comment{Proposal layer}
    \State $\tilde{s}_t^{\mathcal{F}} \leftarrow \mathrm{QuantizedProposal}_{\ell}(X_t)$
      \Comment{approx.\ scoring with INT2 keys}
    \State $\mathcal{C}_t^{\mathcal{F}} \leftarrow \operatorname{TopK}\!\big(\tilde{s}_t^{\mathcal{F}},\, \rho\kappa\big)$
      \Comment{expanded candidate pool, size $n=\rho\kappa$}
    \State $\mathcal{T}_t^{\mathcal{F}} \leftarrow \mathcal{C}_t^{\mathcal{F}}[1{:}\kappa]$
      \Comment{proposal layer attends to its own top-$\kappa$}
  \Else \Comment{Refined layer, $\ell \in \mathcal{S}$}
    \State $\widehat{s}_t^{\mathcal{S}} \leftarrow \mathrm{Refine}_{\ell}\!\big(X_t,\, \mathcal{C}_t^{f(\mathcal{S})}\big)$
      \Comment{rescore only within the inherited pool with INT2 keys}
    \State $\widehat{\mathcal{T}}_t^{\mathcal{S}} \leftarrow \operatorname{TopK}\!\big(\widehat{s}_t^{\mathcal{S}},\, \kappa\big)$
  \EndIf
  \State $ \mathrm{SparseAttn}_{\ell}\!\big(X_t,\, \mathcal{T}_t^{\ell}\big)$
    \Comment{$\mathcal{T}_t^{\ell}\!=\!\mathcal{T}_t^{\mathcal{F}}$ if $\ell\in\mathcal{F}$, else $\widehat{\mathcal{T}}_t^{\mathcal{S}}$}

\EndFor
\end{algorithmic}
\end{algorithm}

\section{Scaling Prism Attention to GLM-5.2}

We further apply Prism Attention to GLM-5.2 (78 layers), evaluating under a unified sparse budget of $\kappa=2048$. Table~\ref{tab:glm_pattern} lists the three evaluated layer-assignment patterns, and Table~\ref{tab:glm_summary} reports the overall accuracy on LongBench-V2 (128K) and RULER (64K/128K); the per-subtask RULER results are further detailed in Table~\ref{tab:glm_ruler_subtask}. Introducing Prism Attention recovers the accuracy that IndexCache loses when its $\mathcal{F}$ layers are reduced, matching the default pattern with far fewer proposal layers.

\begin{table}[t]
\centering
\caption{Layer-assignment patterns evaluated on GLM-5.2 under a unified sparse budget $\kappa=2048$ and proposal expansion factor $\rho=4$. \textbf{Note:} the meanings of F, S, and R here differ from those in Table~\ref{tab:searched_patterns}, and throughout this table follow the definitions given below. \textbf{F} (Full) layers construct sparse indices from the full context, \textbf{S} (Shared) layers directly reuse the indices of their nearest preceding index-producing layer, and \textbf{R} (Refined) layers refine an inherited candidate pool. \emph{21F57S} is the default pattern of GLM-5.2 itself, i.e., its native trained IndexCache configuration, where each F layer builds its own $\kappa$ indices and each S layer reuses these $\kappa$ indices from the nearest preceding F layer. \emph{10F68S} is an IndexCache variant calibrated to fewer F layers. \emph{10F11R57S} incorporates Prism as a hybrid of F, R, and S layers: each F layer builds an expanded pool of $\rho\kappa$ candidate indices, each R layer selects its own top-$\kappa$ from the $\rho\kappa$ pool of the nearest preceding F layer, and each S layer reuses the $\kappa$ indices of its nearest preceding F or R layer.}
\label{tab:glm_pattern}
\renewcommand{\arraystretch}{1.3}
\resizebox{\linewidth}{!}{%
\begin{tabular}{ll}
\toprule
{Config} & {Pattern} \\
\midrule
21F57S (Default) & $\mathcal{FFFSSSFSSSFSSSFSSSFSSSFSSSFSSSFSSSFSSSFSSSFSSSFSSSFSSSFSSSFSSSFSSSFSSSFSSSFSSS}$ \\
10F68S & $\mathcal{FFSSSSSSSSSSSSFSSSFSSSSSSSFSSSSSSSFSSSSSSSFSSSSSSSSSSSSSSSSSSSFSSSSSSSFSSSFSSS}$ \\
10F11R57S & $\mathcal{FFRSSSRSSSRSSSFSSSFSSSRSSSFSSSRSSSFSSSRSSSFSSSRSSSRSSSRSSSRSSSFSSSRSSSFSSSFSSS}$ \\
\bottomrule
\end{tabular}}
\end{table}

\begin{table}[t]
\centering
\caption{Accuracy on GLM-5.2 with the SGLang framework under a unified sparse budget $\kappa=2048$. Each cell reports accuracy and, to the right of the slash, the average output length. The \texttt{\_$L$} suffix in each dataset name denotes the maximum input length.}
\label{tab:glm_summary}
\renewcommand{\arraystretch}{1.2}
\begin{tabular}{lccc}
\toprule
Config & LongBenchV2\_128K & Ruler\_64K & Ruler\_128K \\
\midrule
21F57S              & 65.91 / 4696 & 96.03 / 3802 & 95.53 / 3254 \\
10F68S              & 65.96 / 4972 & 95.68 / 3507 & 93.05 / 3224 \\
10F11R57S           & 66.06 / 4798 & 96.24 / 3086 & 95.15 / 2752 \\
\bottomrule
\end{tabular}
\end{table}

\begin{table*}[t]
\centering
\caption{Per-subtask accuracy (\%) on RULER (64K and 128K) with GLM-5.2 under
the SGLang framework and a unified sparse budget $\kappa=2048$.}
\label{tab:glm_ruler_subtask}
\setlength{\tabcolsep}{4pt}
\renewcommand{\arraystretch}{1.2}
\resizebox{\linewidth}{!}{%
\begin{tabular}{ll ccc ccc cc cc cc c c}
\toprule
& Config
& \rotatebox[origin=c]{30}{S-NIAH-1} & \rotatebox[origin=c]{30}{S-NIAH-2} & \rotatebox[origin=c]{30}{S-NIAH-3}
& \rotatebox[origin=c]{30}{MK-NIAH-1} & \rotatebox[origin=c]{30}{MK-NIAH-2} & \rotatebox[origin=c]{30}{MK-NIAH-3}
& \rotatebox[origin=c]{30}{MQ-NIAH} & \rotatebox[origin=c]{30}{MV-NIAH}
& \rotatebox[origin=c]{30}{HotpotQA} & \rotatebox[origin=c]{30}{SQuAD}
& \rotatebox[origin=c]{30}{CWE} & \rotatebox[origin=c]{30}{FWE}
& \rotatebox[origin=c]{30}{VT} & \textbf{AVG} \\
\midrule
\multirow{3}{*}{\rotatebox[origin=c]{90}{64K}}
& 21F57S    & 100.00 & 100.00 & 98.00  & 100.00 & 100.00 & 100.00 & 99.00  & 100.00 & 79.00 & 84.50 & 97.90 & 90.00 & 100.00 & 96.03 \\
& 10F68S    & 100.00 & 100.00 & 100.00 & 100.00 & 94.00  & 99.00  & 100.00 & 100.00 & 75.00 & 85.75 & 94.70 & 95.33 & 100.00 & 95.68 \\
& 10F11R57S & 100.00 & 100.00 & 100.00 & 99.00  & 100.00 & 100.00 & 97.00  & 100.00 & 78.00 & 82.83 & 99.30 & 95.00 & 100.00 & 96.24 \\
\midrule
\multirow{3}{*}{\rotatebox[origin=c]{90}{128K}}
& 21F57S    & 100.00 & 100.00 & 100.00 & 100.00 & 100.00 & 100.00 & 100.00 & 98.00  & 76.00 & 88.17 & 95.00 & 84.67 & 100.00 & 95.53 \\
& 10F68S    & 100.00 & 100.00 & 100.00 & 100.00 & 86.00  & 88.00  & 96.00  & 98.50  & 76.00 & 84.83 & 89.60 & 90.67 & 100.00 & 93.05 \\
& 10F11R57S & 100.00 & 100.00 & 98.00  & 100.00 & 98.00  & 100.00 & 100.00 & 94.50  & 80.00 & 79.50 & 95.00 & 92.00 & 100.00 & 95.15 \\
\bottomrule
\end{tabular}}
\end{table*}


\clearpage

\section{Limitation and Future Work}
Prism inherits its Proposal/Refined layer partition from the offline greedy calibration procedure of IndexCache: the partition is searched once on a held-out mathematical-reasoning corpus under a token-level cross-entropy objective and then frozen for all downstream evaluations. We adopt this recipe as is and do not treat the calibration procedure itself as a design variable. Since our analysis shows cross-layer index drift to be markedly task-dependent, a systematic exploration of offline calibration along three axes---search strategies stronger than greedy placement, objectives aligned more directly with downstream accuracy, and calibration corpora whose domain and length distribution match the target workload---is likely to yield a more favorable layer assignment at no additional inference cost.


\end{document}